%% file: main.tex
\documentclass[nohyperref]{article}

\usepackage{microtype}
\usepackage{graphicx}
\usepackage{subfigure}
\usepackage{booktabs} % for professional tables
\usepackage{wrapfig}
\usepackage{hyperref}

\usepackage[accepted]{icml2022}
\usepackage{enumitem}
\usepackage{amsmath}
\usepackage{amssymb}
\usepackage{mathtools}
\usepackage{amsthm}
\usepackage[normalem]{ulem}
\usepackage{multirow}
\usepackage{amsfonts}
\usepackage{bm}
\usepackage{xcolor,colortbl}

\usepackage{eqparbox}
\usepackage{makecell}
\usepackage{threeparttable}

\renewcommand{\algorithmiccomment}[1]{\hfill{\scriptsize $\rhd$ #1}}
\usepackage[font=small,labelfont=it]{caption}
\usepackage[capitalize,noabbrev,nameinlink]{cleveref}

\theoremstyle{plain}

\theoremstyle{definition}

\theoremstyle{remark}

\usepackage[textsize=tiny]{todonotes}

\newcommand{\eat}[1]{}

\definecolor{Red}{rgb}{1,0,0}
\definecolor{Blue}{rgb}{0,0,0.8}
\definecolor{Green}{rgb}{0,0.4,0.7}
\definecolor{airforceblue}{rgb}{0.36, 0.54, 0.66}
\definecolor{ao(english)}{rgb}{0.0, 0.5, 0.0}
\definecolor{azure(colorwheel)}{rgb}{0.0, 0.5, 1.0}
\definecolor{crimson}{rgb}{0.86, 0.08, 0.24}
\definecolor{darkcerulean}{rgb}{0.03, 0.27, 0.49}
\definecolor{cobalt}{rgb}{0.0, 0.28, 0.67}
\definecolor{rosegold}{rgb}{0.72, 0.43, 0.47}
\definecolor{orange-red}{rgb}{1.0, 0.27, 0.0}
\definecolor{mountainmeadow}{rgb}{0.19, 0.73, 0.56}
\definecolor{malachite}{rgb}{0.04, 0.85, 0.32}
\definecolor{darkblue}{rgb}{0.0, 0.0, 0.55}

\definecolor{customblue}{rgb}{0.2, 0.35, 0.8}

\usepackage{hyperref}
\hypersetup{colorlinks=true}
\hypersetup{linktoc=all}
\hypersetup{citecolor=customblue}
\hypersetup{linkcolor=crimson}
\hypersetup{urlcolor=MidnightBlue}
\usepackage[all]{hypcap}
\usepackage[percent]{overpic}
\usepackage[usestackEOL]{stackengine}

\usepackage[nameinlink]{cleveref}
\creflabelformat{equation}{#2\textup{#1}#3}  
\crefname{assumption}{assumption}{assumptions}

\definecolor{gg}{gray}{0.9}
\newcolumntype{a}{>{\columncolor{gg}}c}

\newcommand{\bsy}{\boldsymbol}

\creflabelformat{equation}{#2\textup{#1}#3}  

\definecolor{plRed}{rgb}{0.8,0,0}
\definecolor{miBlue}{rgb}{0,0,0.8}
\newcommand{\plhi}[1]{{\color{plRed}{#1}}}
\newcommand{\mihi}[1]{{\color{miBlue}{#1}}}

\icmltitlerunning{Bitwidth Heterogeneous Federated Learning with Progressive Weight Dequantization}

\begin{document}

\twocolumn[
\icmltitle{Bitwidth Heterogeneous Federated Learning with\\ Progressive Weight Dequantization}


\icmlsetsymbol{equal}{*}

\begin{icmlauthorlist}
\icmlauthor{Jaehong Yoon}{kaist,equal}
\icmlauthor{Geon Park}{kaist,equal}
\icmlauthor{Wonyong Jeong}{kaist}
\icmlauthor{Sung Ju Hwang}{kaist,aitrics}
\end{icmlauthorlist}

\icmlaffiliation{kaist}{Korea Advanced Institute of Science and Technology (KAIST), South Korea}
\icmlaffiliation{aitrics}{AITRICS, South Korea}

\icmlcorrespondingauthor{Jaehong Yoon}{jaehong.yoon@kaist.ac.kr}
\icmlcorrespondingauthor{Geon Park}{geon.park@kaist.ac.kr}
\icmlcorrespondingauthor{Sung Ju Hwang}{sjhwang82@kaist.ac.kr}

\icmlkeywords{Machine Learning, ICML}

\vskip 0.3in
]
\printAffiliationsAndNotice{\icmlEqualContribution} % otherwise use the standard text.

\input{contents/0_abstract}

\input{contents/1_intro}

\input{contents/2_related_work}
\input{contents/3_approach_base}
\input{contents/3_approach_ours}

\input{contents/4_experiments}
\input{contents/5_conclusion}

%\bibliography{reference}
%\bibliographystyle{iclr2022_conference}

\bibliography{main}
\bibliographystyle{icml2022}
\input{contents/A_appendix}

\end{document}

%% file: contents/0_abstract.tex
\begin{abstract}
In practical federated learning scenarios, the participating devices may have different bitwidths for computation and memory storage by design. However, despite the progress made in device-heterogeneous federated learning scenarios, the heterogeneity in the bitwidth specifications in the hardware has been mostly overlooked. We introduce a pragmatic FL scenario with bitwidth heterogeneity across the participating devices, dubbed as Bitwidth Heterogeneous Federated Learning (BHFL). BHFL brings in a new challenge, that the aggregation of model parameters with different bitwidths could result in severe performance degeneration, especially for high-bitwidth models. To tackle this problem, we propose ProWD framework, which has a trainable weight dequantizer at the central server that progressively reconstructs the low-bitwidth weights into higher bitwidth weights, and finally into full-precision weights. ProWD further selectively aggregates the model parameters to maximize the compatibility across bit-heterogeneous weights. We validate ProWD against relevant FL baselines on the benchmark datasets, using clients with varying bitwidths. Our ProWD largely outperforms the baseline FL algorithms as well as naive approaches (e.g. grouped averaging) under the proposed BHFL scenario.
%We validate ProWD against relevant FL baselines on multiple benchmarks, using clients with varying bitwidths, and show that ProWD largely ouperforms the base FL algorithms as well as naive approaches for tackling bitwidth heterogeneity, such as grouped averaging.
\end{abstract}

%% file: contents/1_intro.tex
\section{Introduction}
In recent decades, the drastic evolution of hardware technologies for edge devices has changed our lives from the root. Smart edge devices, which have the ability to process sensory inputs and communicate, such as embedded sensors, drones, phones, smart watches, and augmented reality glasses, are now being used in our everyday lives. Such accessibility of edge devices has led to the emergence of \emph{Federated Learning} (FL)~\cite{McMahan2017, zhao2018federated, chen2019communication}, a learning framework in which multiple clients collaboratively train on private local data while periodically communicating the trained models across them, often through a server which aggregates and broadcasts the local models. Many previous works have investigated the potential and applicability of FL in various learning frameworks, such as semi-supervised learning~\cite{jeong2020federated}, bayesian learning~\cite{wang2020federated}, graph neural networks~\cite{mei2019sgnn,wu2021fedgnn}, meta-learning~\cite{jiang2019improving,fallah2020personalized}, and continual learning~\cite{yoon2021federated}. 

A crucial challenge in FL is that there could be a large discrepancy among participants, in their data distribution, tasks, model architectures, and devices which often leads to incompatibility of the models that are being aggregated. This problem is often referred to as Heterogeneous Federated Learning~\cite{jiang2020federated, lin2020ensemble, he2020group, diao2020heterofl} problem. Many existing works have successfully alleviated the adverse effect of data-, model-, and device-heterogeneity. However, the most basic assumption even in such heterogeneous FL scenarios, is that all models have the same bitwidths.

Yet, in real-world FL scenarios, participating devices may have heterogeneous bitwidth specifications. Suppose that we are building a federated network of various health care providers, such as hospitals, community health centers, and clinics, as well as end-users, and even wearable devices, i.e. smart watches. Each client has a health disorder prediction model for the users themselves or their patients that continuously learns to perform a diagnosis given the heart rates or bioelectric signals. The scenario allows the participation of local clients with various hardware infrastructures, some of them using models built under lightweight devices based on low-bitwidth hardware operations using FPGA, ASIC, Raspberry Pi, or Edge GPUs. Here, BHFL enables local devices with different hardware specifications to participate in a single federated learning framework without the need for uniformity of the infrastructure, enhancing the pool of devices that could participate in collaborative learning. We refer to this practical FL scenario as Bitwidth Heterogeneous Federated Learning (BHFL), which we illustrate in~\Cref{fig:bhfl}.
\input{contents/tables/1_compariton}
\input{contents/figures/1_bhfl}

Tackling BHFL is a nontrivial problem, as it poses new challenges such as (i) suboptimal loss convergence due to distributional shift after aggregating the weights of bitwidth-heterogeneous models, and (ii) inherent limitation of expressive power in low-bitwidth weights. As shown in \Cref{tab:categorization}, due to these challenges, existing methods cannot appropriately handle the new setting. While there exists a line of works that propose to quantize model weights to reduce the communication cost when transmitting them to the server, they assume that full-precision weights are being used at the local devices. However, processing full-precision weights may not be possible for resource-limited hardware devices. 

%As shown in \Cref{fig:c1}, a quantization function projects the full-precision parameters into a weight subspace for low-bitwidth clients. While the server fairly sends identical weights to clients, the local updates depend on the limitation of clients' bitwidth, resulting in a substantial weight disparity. This leads a federated learning model to suffer from poor convergence of local clients. We named the problem to weight space discrepancy. Also, since received parameters from low bitwidth clients are composed of fixed quantized values, naive aggregation of different bitwidth parameters makes the weights biased to quantized values from lower bitwidth clients. Consistent aggregation and re-distribution of such biased weights at each communication round is extremely detrimental to the federated learning model and results in poor convergence, as described in \Cref{fig:c2}.}

In this paper, we propose a novel framework that can succesfully deal with the new challenges posed by the BHFL problem, which we name as
{\textbf{Pro}gressive \textbf{W}eight \textbf{D}equantization (\textbf{ProWD})},
The ProWD framework enhances the compatibility across bit-heterogeneous weights with selective weight aggregation and weight dequantization. The selective weight aggregation discards outliers from the low-precision weights, and the trainable dequantizer at the server recovers high-precision weight information from the given low-bitwidth weights. These two methods collaboratively alleviate the information loss resulting from the aggregation of incompatible weights with heterogeneous bitwidths. 

We evaluate the performance of our ProWD on various benchmark datasets and show that our method significantly outperforms previous FL methods that consider weight quantization, while also outperforming naive heuristics to tackle the bitwidth heterogeneity, such as grouped averaging. We also provide comprehensive analyses which show that existing FL methods suffer from poor convergence and adaptation under the bit-heterogeneous federated learning scenario, while ProWD consistently increases the performance of all local models regardless of their bitwidths. 

In summary, our contributions are threefold:
% \vspace{-0.1in}
\begin{itemize}
    \item We define a practical federated learning scenario where the participating devices may have largely different bitwidths, which brings in new challenges such as degenerate distributional shift of federated weights and limited expressive power of low-bitwidth models.
    
    \item We propose \emph{ProWD}, a novel framework that effectively tackles the bitwidth heterogeneous FL problem, by selectively aggregating the weights and hierarchically dequantizing the weights prior to aggregation.
    
    \item We demonstrate the efficacy of our ProWD framework by validating it on diverse compositions of bitwidth specifications in local clients, against recent FL methods as well as naive remedies.
\end{itemize}

%% file: contents/tables/1_compariton.tex
\begin{table*}
\centering
\resizebox{0.9\textwidth}{!}{
\begin{threeparttable}
\small
\caption{\small Categorization of existing methods for bitwidth heterogeneous federated learning.\vspace{-0.05in}}\label{tab:categorization}
\begin{tabular}{lccccccc}
\toprule
\textsc{Methods} &
{FL Type} &
{$\text{Bits}_{\text{Server}}$} & 
{$\text{Bits}_{\text{Clients}}$} &
{$\text{Bits}_{\text{Uplink}}$} & 
{$\text{Bits}_{\text{Downlink}}$} & 
{Communication} \\
\midrule
\textsc{FedAvg}~\cite{McMahan2017} & 
FL & Float32 & Float32 & Float32 & Float32 & Weights \\
\textsc{FedProx}~\cite{li2018federated} &
FL & Float32 & Float32 & Float32 & Float32 & Weights \\
\textsc{FedPAQ}~\cite{reisizadeh_fedpaq_2020} &
~~QPC$^1$ & Float32 & Float32 & Target bits & Float32 & ~$Q_\text{Target\_bits}$(Diffs)$^2$ \\
\textsc{FedCOM}~\cite{haddadpour_federated_2020} &
QPC & Float32 & Float32 & Target bits & Float32 & $Q_\text{Target\_bits}$(Diffs) \\
\textsc{FedCOMGATE}~\cite{haddadpour_federated_2020} &
QPC & Float32 & Float32 & Target bits & Float32$\times$2 & $Q_\text{Target\_bits}$(Diffs) \\
\textsc{ProWD (Ours)} &
BHFL& Float32 & Client-specific & Client-specific & Client-specific & $W_Q$$^3$  \\
\bottomrule
\end{tabular}
\begin{tablenotes}
\item [1] Quantized Parameter Communication (QPC) ~~~~~~$~~~~~~~~~~~~~~~~~~~~~~~~^2$~$Q_\text{Target\_bits}(w)$: A function quantizes input $w$ to the target bitwidth
%\item [2] $Q_\text{Target\_bits}(w)$: A function quantizes input $w$ to a bitwidth $s$ $~~~~^3$asd
\item [3] Weights obtained from clients' bit-dependent training
\end{tablenotes}
\end{threeparttable}}
\end{table*}

%% file: contents/figures/1_bhfl.tex
%\begin{wrapfigure}{R}{0.45\textwidth}
\begin{figure}
\centering
% \vspace{-0.05in}
\includegraphics[width=0.45\textwidth]{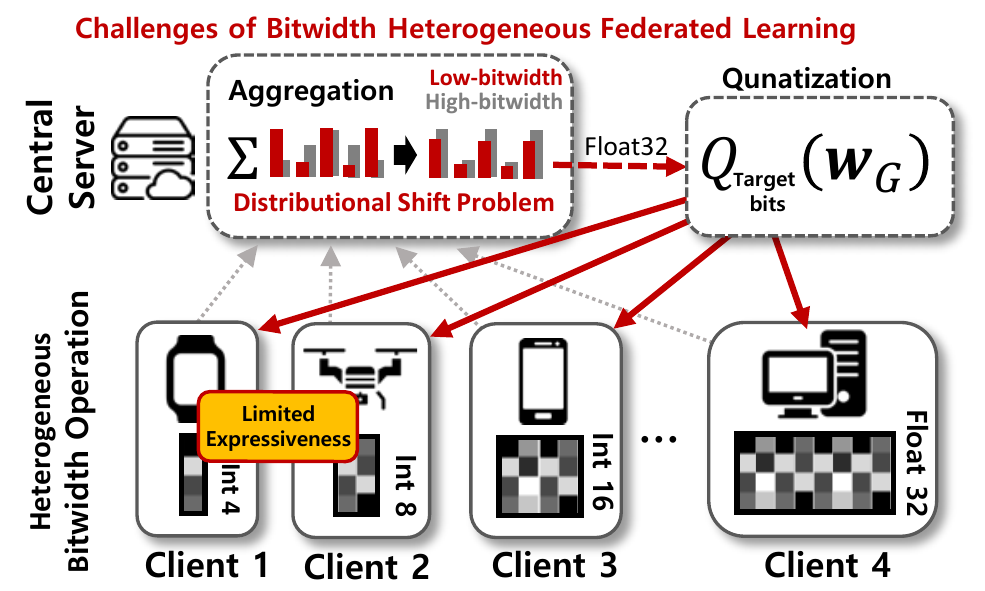}
% \vspace{-0.1in}
\caption{\label{fig:bhfl} \small \textbf{Bitwidth Heterogeneous Federated Learning.} We consider a FL setting where the participating devices have heterogeneous bitwidths. FL with different bitwidth models may cause detrimental side effects due to (i) distributional shift of model weights, and (ii) the limited expressiveness in low-bit weights.}
% \vspace{-0.1in}
\end{figure}
%\end{wrapfigure}

%% file: contents/2_related_work.tex
\section{Related Works}

\paragraph{Quantization for Federated learning}
While no existing work considers the problem of federated learning across devices with different innate bitwidths, several works propose to quantize weights or gradients to reduce the communication cost, often referred to as Quantized Parameter Communication (QPC). FedPAQ~\cite{reisizadeh_fedpaq_2020} proposed to send the quantized changes of the weights at each client, instead of the full weights.  FedCOMGATE~\cite{haddadpour_federated_2020} extends this idea by adopting a global learning rate and guiding the update direction to stay close to each other with an accumulated correction vector, to further address data heterogeneity. While we empirically observed that QPC approaches alleviate the performance degeneration of the high-bitwidth models under the BHFL scenario since they do not drastically change the weights as simple averaging does, they are suboptimal since they do not explicitly tackle the challenges posed by the BHFL problem.

% While no existing work considers the problem of federated learning across devices with different innate bitwidths, several works propose to quantize weights or gradients to reduce the communication cost, often referred to as Quantized Parameter Communication (QPC). FedPAQ~\cite{reisizadeh_fedpaq_2020} proposed to send the quantized changes of the weights at each client, instead of the full weights.  FedCOMGATE~\cite{haddadpour_federated_2020} extends this idea by adopting a global learning rate and guiding the update direction to stay close to each other with an accumulated correction vector, to further address data heterogeneity. While we empirically observed that QPC approaches alleviate the performance degeneration of the high-bitwidth models under the BHFL scenario since they do not drastically change the weights as simple averaging does, they are suboptimal since they do not explicitly tackle the challenges posed by the BHFL problem. 

% \vspace{-0.05in}
\paragraph{Low-bitwidth training} 

Low-bitwidth training is an approach to train a model using only low-bitwidth operations and data types at training time, which enables on-device learning with lightweight edge devices. BNN~\cite{hubara_binarized_2016} and XNOR-Net~\cite{rastegari_xnor-net_2016} are approaches to accelerate the training of the convolutional neural networks using binary operations in the forward pass with binarized weights and activations, with floating-point operations to compute gradients in the backward pass. Except for that, diverse approaches have been proposed which focus on using 8-bit floating point numbers for the weights, activations, and gradients~\cite{zhou_dorefa-net_2018}, devising ternary gradients to reduce communication cost (yet, it allows float32 operation for accumulating the gradients)~\cite{wen_terngrad_2017}, training without floating-point operations~\cite{wu_training_2018}, adopting direction-sensitive gradient clipping~\cite{zhu_towards_2020}, training for mixed-bitwidth models~\cite{zhang_fixed-point_2020,zhao_distribution_2021,sun_ultra-low_2020}. Most existing neural quantization approaches quantize only the weights or the gradients, while preserving parts of the network that has a large impact on the performance, such as activations, in full-precision, and thus are not applicable to training on devices with limited bitwidths. 

\paragraph{Neural dequantization} 
The dequantization of low-bitwidth signals into high-bitwidth signals has been studied for diverse applications. For image reconstruction, \citet{xing2021invertible} aims to recover the resolution of quantized sRGB images to full-dynamic-range RAW image data via an invertible dequantizer function. In generative flow models, \citet{nielsen_closing_2020} dequantizes the discrete-valued data by adding a uniform noise to guarantee that the data is able to have any value in the continuous domain. Dequantizatoin is also used for the process of converting the low-bit representation of the weights to high-bit without changing their values~\cite{gholami_survey_2021}, or to decode the encoded vectors using a codebook~\cite{lee_flexor_2020}.
In this paper, we refer to ``dequantization'' to describe the process of recovering the original high-bitwidth weights from the low-bitwidth quantized weights received from local clients , so that a server can utilize the high-performing recovered models for aggregation during FL.
To our knowledge, our work is the first work that provides an appropriate method adopting the weight dequantization approach for solving practical FL scenarios.

%\citet{xing2021invertible} allows high-quality recovery of full-dynamic-range RAW image data from quantized sRGB images by using an invertible dequantizer network. In flow models, dequantization is used for the discrete data to enforce continuity of the input domain~\cite{nielsen_closing_2020}. In the context of neural network weights, the term ``dequantization'' is used in existing literature for converting the low-bit representation of the weights (i.e. Int8) to a high-bit representation (i.e. Float32) without changing the values~\cite{gholami_survey_2021}, or in case of vector quantization, decoding the encoded vectors using a codebook~\cite{lee_flexor_2020}. In contrast, what we refer to as ``dequantization'' in this paper is recovering the original high-precision weight values from low-bit quantized weight values. Ideally, we want higher model accuracy after passing the weights through the dequantizer. We utilize this weight dequantizer in the federated learning setting for recovering the high-precision weights from quantized low-precision weights sent from the client. To the best of our knowledge, our work is the first direction that provides an appropriate method adopting the dequantization approach for solving practical federated learning scenarios.

%% file: contents/3_approach_base.tex
\section{Bitwidth Heterogeneous Federated Learning (BHFL)}
We now introduce the problem setup for standard federated learning and extends it to the Bitwidth Heterogeneous Federated Learning (BHFL) scenario, where a subset of the participating devices train the local models with low-bitwidth operations, according to their hardware specifications (\Cref{subsec:setup}). We then describe the quantized computational flow for the training of low-bitwidth clients in \Cref{subsec:wage}.

\subsection{Problem statements}\label{subsec:setup}
\paragraph{Federated Learning} In a standard Federated Learning (FL) scenario~\citep{McMahan2017, chen2019communication}, each client trains the local model on the private data and periodically %transmits learned model to the server to integrate multiple local knowledge and preserve the data privacy. 
transmits the model parameters to the central server, where the models are aggregated and broadcasted back to the clients.
Let $N$ different clients $\mathcal{C}=\left\{c_1, ..., c_N\right\}$ participate in an FL system. Given training samples $\bm{x}_n$ and its corresponding labels $\bm{y}_n$, we suppose that a client $c_n$ solves a local optimization problem $\mathcal{L}_n=\text{CE}(f(\bm{x}_n;\bm{w}_n);\bm{y}_n)$, where CE is a cross-entropy loss and $f(\cdot;\bm{w}_n)$ is a neural network of client $c_n$ parameterized by $\bm{w}_n$. At each communication round $r$, clients $\mathcal{C}^{(r)}\subseteq\mathcal{C}$ send the model parameters to the central server, and the server aggregates received weights, for exampling by averaging their weights.% as follows: $\bm{w}_G^{(r)} = \frac{1}{|\mathcal{C}^{(r)}|}\sum_{n}\bm{w}^{(r)}_n$ for $c_n\in\mathcal{C}^{(r)}, \forall n$, then re-distributes it to the local clients to update local weights, $\bm{w}^{(r+1)}_n\leftarrow\bm{w}_G^{(r)}$. 
%The goal of Centralized Federated Learning is to obtain the best-performing collaborative model $\bm{w}_G$ for all local tasks. In contrast, Personalized Federated Learning focuses on the adaptation of its local models $\{\bm{w}_n\}^N_{n=1}$ by leveraging the learned knowledge from other clients. 
% \vspace{-0.1in}
\paragraph{Bitwidth heterogeneous federated learning} 
%In bit-heterogeneous federated learning, we break the standard assumption that all clients are capable of efficient full-precision floating-point operations. Instead, we assume that the set of $N$ clients $\left\{c^f_1, \dots, c^f_{N_f}, c^q_1, \dots, c^q_{N_q} \right\}$ consist of $N_f$ \textbf{full-precision clients} capable of high computational load and $N_q$ \textbf{quantized clients} which have limited computational capacity.  For the quantized clients, we adopt the quantized training method proposed in \cite{wu_training_2018} in order to locally train a neural network entirely using low bitwidth integer operations. Each quantized client $c^{q}_i$ exclusively uses $b_i$-bit integer operations while training and evaluating the network.

The majority of existing FL methods assume full-precision operations for local clients, even when they consider device heterogeneity. However, the participating devices have largely heterogeneous bitwidths according to their hardware specifications. To this end, we introduce a practical FL scenario, named Bitwidth Heterogeneous Federated Learning (BHFL), in which we relax the strong assumption that all clients are capable of full-precision floating-point operations. We represent $n$-th local client as a tuple of the model weights $\bm{w}_n$ and the corresponding hardware bitwidth information $s_n\in\mathcal{S}$, where %$\mathcal{S}=\{$Int6 Int8, Int16, Float32$\}$.
$\mathcal{S}$ is a set of available bitwidth specifications for local hardware devices. 
Then, the set of $N$ clients, $\mathcal{C}$ can be represented as follows: $\mathcal{C}=\left\{(\bm{w}_1, s_1), \dots, (\bm{w}_N, s_N)\right\}$. At each round of communication, a central server receives client tuples and aggregates model parameters which might be quantized in various levels depending on hardware specifications. 
%Since the aggregated parameters are full-precision, the server redistributes them to connected clients after quantizing with respect to their hardware specification: $\bm{w}_i\leftarrow Q(\bm{w}_G,s_n)$, where $i$ is an index of the client.
We assume that the server allows the full-precision computation, which redistributes $\bm{w}_G^{(r)}$ to all clients after quantizing the aggregated weights according to the hardware specifications for each client; that is, $\bm{w}_n^{(r+1)}\leftarrow Q(\bm{w}_G^{(r)},s_n), \forall n$. %for $n^{th}$ client.

\input{contents/figures/1_challenges}
In BHFL, what aggregation method we use could have a strong impact on the overall performance of the model being learned. Using a naive averaging technique, such as simple averaging, to aggregate local clients' model parameters hinders the convergence. This is because the model falls into a suboptimal local minimum due to the incompatibility across the model weights with heterogeneous bitwidths, due to large discrepancy in their distributions. The detrimental effect is more severe for larger bitwidth models, as aggregating the low-bitwidth model weights will result in the loss of expressiveness in the model. In \Cref{fig:challenge}, we observe that the distribution of the full-precision weights quickly degenerates into three peaks that correspond to the ternary quantization values of the low-bitwidth models.
%\TBD{This is because the model weights are transmitted as ternary values for low-bitwidth uplink transmission.}

\input{contents/figures/3_method}

%\subsection{Quantized Training for Limited Bitwidth Clients}\label{subsec:wage}
\subsection{Local Computation for Limited Bitwidth Clients}\label{subsec:wage}
We assume that the local clients are edge devices that perform limited bitwidth computations according to their hardware specifications. Float32 clients perform regular full-precision training, but quantized clients, such as Int6, Int8, and Int16, perform training using low-precision integer operations, following \citet{wu_training_2018}.

%\paragraph{Forward pass.} %Let $\bm{w}_f^l$ be the weights in the $l$-th layer of a full-precision client. The quantized $s_n$-bit weights that correspond to $\bm{w}_f^l$ is $\bm{w}_q^l$. Let us denote the convolution operator with $*$. Since convolution involves multiplication between the weights and the activations, naively computing $\bm{w}_q^l * \bm{a}_q$ requires multiplication of two $s_n$-bit integers, as well as more than $(2\cdot s_n)$-bit integers to store the intermediate result without truncation. To alleviate this problem, we first ternarize the quantized weights with the function $Q_{I2}(\cdot)$, and then compute the convolution. This way, only $s_n$-bit addition and subtraction operations are required, and the intermediate result can be stored in $(s_n+1)$ bits.
Let $\bm{q}^{l}$ and $\bm{a}^{l}$ be $s$-bit quantized weights and activations at layer $l$ for a client, respectively. We denote the convolution operator as $*$. Since convolution involves multiplication between the weights and the activations, naively computing $\bm{q}^l * \bm{a}^{l-1}$ requires the hardware to support fast multiplication of two $s$-bit integers. To relax this requirement, we ternarize the quantized weights before the convolution operation:%
%To avoid this problem, we first ternarize the quantized weights with the function $Q_{\text{Int2}}(\cdot)$, and then compute the convolution. Thus, only $s_n$-bit addition and subtraction operations are required, and the intermediate result can be stored in $(s_n+1)$ bits.
%After performing convolution, we scale down the result by a power-of-two coefficient $\alpha^l$ defined for each layer. How $\alpha^l$ is determined is elaborated in the following sections. The reason for scaling the activations is because the scale of the weight initialization is much bigger than the full-precision models, so we need to compensate for the difference. After that, we ensure that the result is in the allowed range for $s_n$-bit integers by applying the quantizer function $Q_\text{s}(\cdot)$. Finally, we apply the nonlinearity $\text{ReLU}(\cdot)$:
%\eat{We further introduce a power-of-two coefficient $\alpha^l$ per layer that scales down the results. After performing convolution, we scale down the result by a power-of-two coefficient $\alpha^l$ defined for each layer, since scale of the weight initialization is much bigger than the full-precision models, so we need to compensate for the difference. After that, we ensure that the result is in the allowed range for $s_n$-bit integers by applying the quantization function $Q_\text{s}(\cdot)$:}
\abovedisplayskip=12pt plus 0pt minus 3pt
\belowdisplayskip=12pt plus 0pt minus 3pt
\begin{align}\label{eq:af}
\bm{a}^{l} = \text{ReLU}\left(Q_{s}\left(Q_\text{Int2}\left(\bm{q}^l\right) * \bm{a}^{l-1} / \alpha^l\right)\right).
%\bm{a}^{l}_q &= \text{ReLU}\left(Q_{s}\left(Q_{I2}\left(\bm{w}^l_q\right) * \bm{a}^{l-1}_q / \alpha^l\right)\right),
\end{align}
%\TBD{In this way, only $s$-bit addition and subtraction operations are required in both training and inference time, and the intermediate results can be stored in $s+1$ bits, instead of $2\cdot s$ bits. Also, compared to binarization, we can gain much more expressiveness, thus achieving a good tradeoff between precision and computational requirements~\cite{li_ternary_2016}.} 
Thus, \Cref{eq:af} only needs $s$-bit addition and subtraction operations for training and inference, and the intermediate results can be stored in $(s+1)$-bits while achieving a good tradeoff between precision and computation~\cite{li_ternary_2016}.
To prevent the activations from clipping out of range, we rescale the activations with a pre-defined layerwise scalar coefficient $\alpha^l$ so that they are within the range for $s$-bit integers. This rescaling can be implemented efficiently with a hardware bit-shift operation. We further provide details for training limited-bitwidth clients in \Cref{appendix:low_bit_traing}. After a few local training steps, limited-bitwidth clients broadcast original low-bitwidth weights $\bm{q}$ or the ternarized weights $Q_\text{Int2}\left(\bm{q}\right)$ to the server, like other QPC methods~\cite{reisizadeh_fedpaq_2020,haddadpour_federated_2020}.

% The $s$-bit quantizer funtcion $Q_s(\cdot)$ is defined elementwise as follows:
% \begin{align*}
%     Q_s(x) &= \mathrm{clip}_s\left(\left\lceil 2^{s-1} \cdot x \right\rfloor / 2^{s-1}\right), \\
%     \mathrm{clip}_s(x) &= \mathrm{clip}\{x, -1 + 2^{1-s}, 1 - 2^{1-s}\}.
% \end{align*}

%% file: contents/figures/1_challenges.tex
\begin{figure}
\centering
% \vspace{-0.05in}
\includegraphics[width=0.4\textwidth,trim=0.2cm 0cm 0.3cm 0cm,clip]{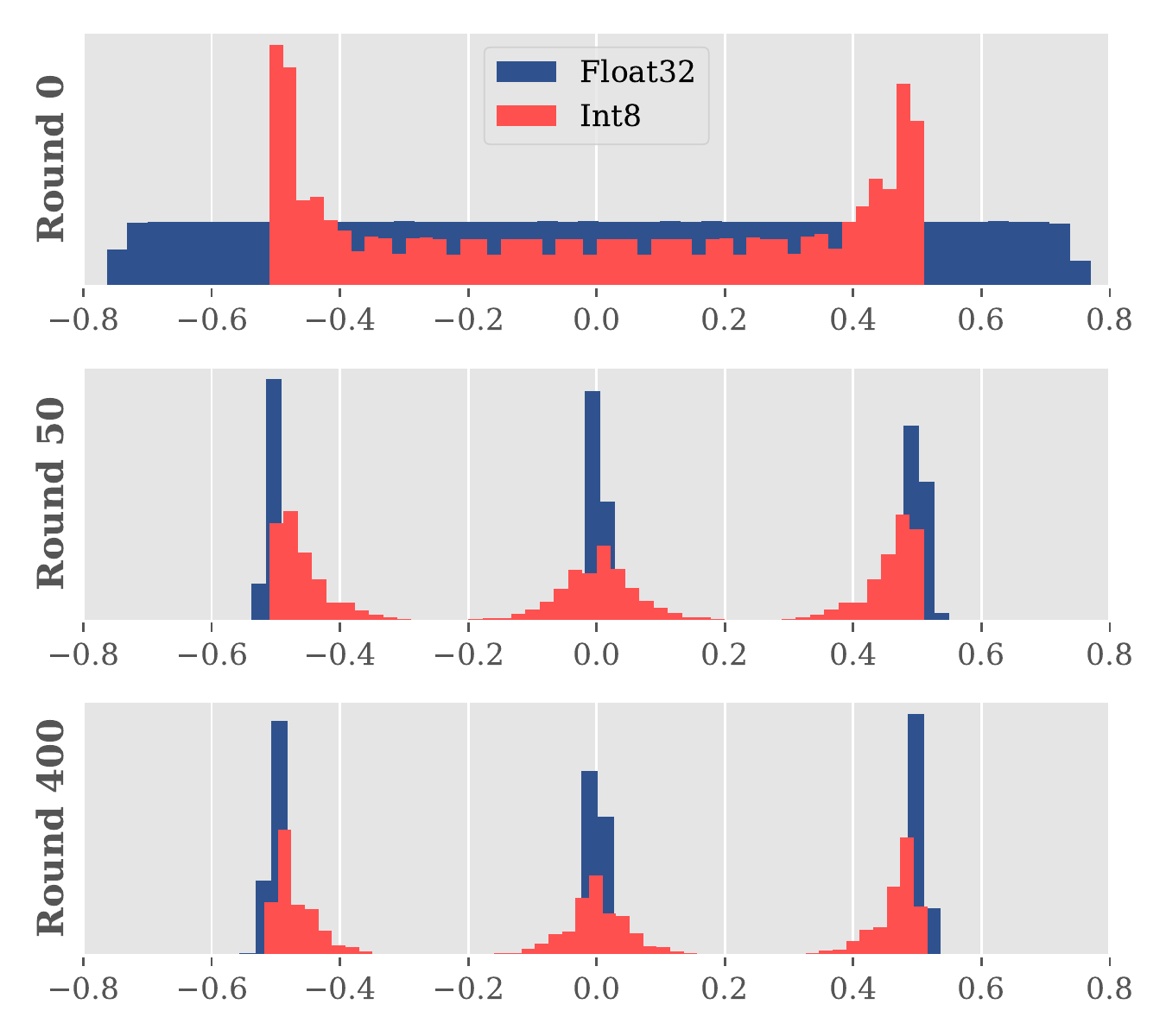}
% \vspace{-0.15in}
\caption{\label{fig:challenge} \small 
\textbf{Skewed weight distribution after the aggregation of mixed bitwidth weights}.
The distribution of the last layer's weights of the full-precision and low-bitwidth models at the initial, after $50$, and $400$ aggregation rounds.}
% \vspace{-0.05in}
\end{figure}

%% file: contents/figures/3_method.tex
\begin{figure*}[t!]
    % \vspace{-0.05in}
    \centering
    \small
    \begin{tabular}{cc}
    \includegraphics[height=3.2cm]{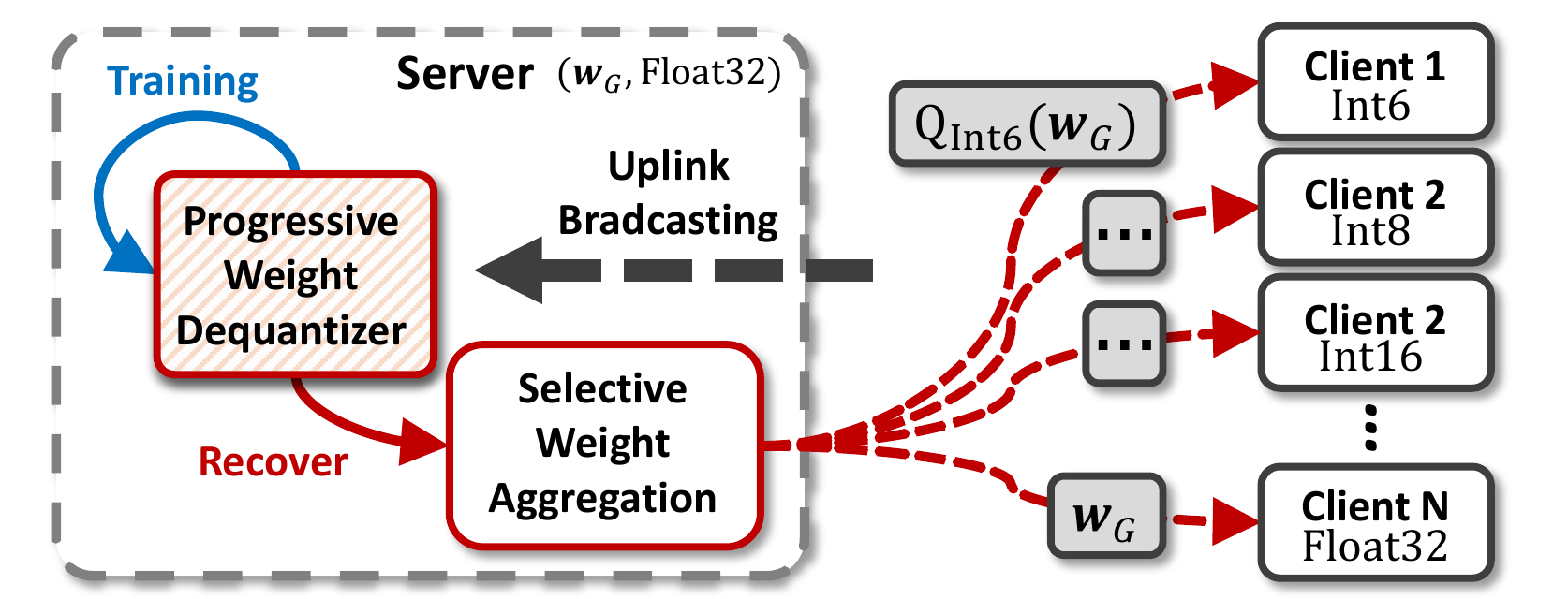}&\hspace{-0.25in}
    \includegraphics[height=3.2cm]{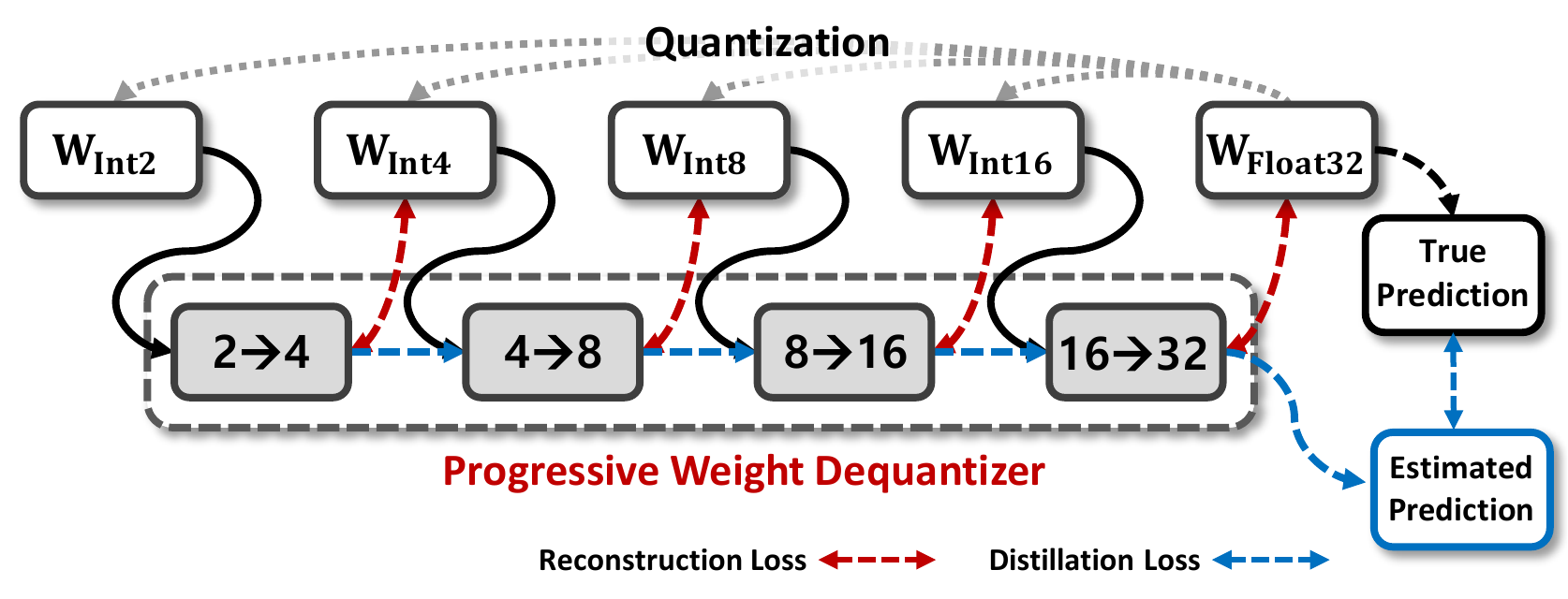}\\
    (a) Bitwidth heterogenrous federated learning with ProWD&\hspace{-0.25in}
    (b) An example of progressive weight dequantizer
    \end{tabular}
    % \vspace{-0.1in}
    \caption{\footnotesize \textbf{(a) Illustration of our ProWD framework.} Clients send their local models and hardware bitwith specifications to the server. We reduce the distributional disparity among weights from different bitwidth devices during BHFL by introducing weight dequantization and selective aggregation. \textbf{(b) Progressive weight dequantizer} recovers low-bit weights into the high-bit via minimizing two loss terms.}
    \label{fig:method}
    % \vspace{-0.15in}
\end{figure*}

%% file: contents/3_approach_ours.tex
\section{Bitwidth Heterogeneous FL with ProWD}\label{sec:ProWD}
%Bitwidth Heterogeneous Federated Learning (BHFL) poses critical challenges: (i) biased weight distribution after the aggregation of mixed bitwidth weights, (ii) inherently limited expressive power of low-bitwidth weights. 
We first introduce our naive remedies for alleviating the challenges posed by the BHFL problem, and their limitations in \Cref{subsec:naive}. Then, we propose a novel FL framework to properly handle BHFL scenarios, named as \emph{ProWD}, which consists of two core components: progressive weight dequantization and score-based selective weight aggregation, described in \Cref{subsec:deq} and \Cref{subsec:sgu}, respectively.

\subsection{Naive Remedies for Bitwidth Heterogeneity in FL}\label{subsec:naive}
BHFL deteriorates the performance of higher-bitwidth clients' models, while the local models from low-bitwidth clients often enjoy clear benefits due to knowledge transfer from the high-bitwidth models (See \Cref{fig:challenge}). We first suggest a bitwidth-dependent averaging technique for BHFL, preventing interference across different bitwidth weights during aggregation, referred to as FedGroupedAvg. However, this method suffers from poor transferability, as the rich knowledge obtained by expressive high-bitwidth models is not transferred to low-bitwidth ones, while it preserves the performance of higher-bitwidth clients. Thus, we additionally suggest a modified version of FedGroupedAvg which allows the knowledge transfer from higher-bitwidth clients to the lower ones, but not vice versa, named as FedGroupedAvg-Asymmetric. These two simple baselines can improve the performance of local model, but are suboptimal in that it does not utilize the full knowledge of the participating models. Thus, we propose a novel method to effectively tackle the challenges in BHFL, which overcomes the limitations of such naive remedies.

\subsection{Progressive Weight Dequantization}\label{subsec:deq}
%While the server selectively utilizes the received weights to prevent the deteriorating shift in distributions and encourages informative knowledge aggregation among different bitwidth weights, 
Low-bitwidth models, while hardware-friendly, severely limits the expressiveness of the model. Aggregating the parameters with such limited information thus may degrade the quality of high-bitwidth models during federated learning. To this end, we propose a trainable dequantizer that reconstructs the full-precision weights for given low-precision weights. 
Yet, directly reconstructing the low-bit weights to the high-bit may not be as effective, especially when the bitwidth gap is large, since there exists a significant disparity between their distributions. To alleviate this issue with single-step reconstruction, our dequantizer breaks down the problem into block-wise weight dequantization problems by adopting a stack of network blocks. 

%Let a set of bitwidth precisions $\Pi=\{\pi_0,\cdots,\pi_k\}$ be a non-strict superset of $\mathcal{S}$ (i.e., $\mathcal{S}\subseteq\Pi$), where the bitwidth $\pi_j\in\Pi$ is lower than $\pi_{j+1}$, $\forall j$, and \modify{the lowest and highest bitwidths are equal in each set, $\pi_0=s_0$ and $\pi_{k}=s_m$}.
Let a $\Pi=\{\pi_0,\cdots,\pi_k\} \supseteq \mathcal{S}$ be an ordered set of bitwidth precisions, where $\pi_j$ is a lower precision bitwidth than $\pi_{j+1}$ $\forall j$. 
The lowest and the higest bitwidths of $\Pi$ correspond to those of  $\mathcal{S}$, i.e., $\pi_0=s_1$ and $\pi_{k}=s_m$.
We denote a progressive weight dequantization function $\phi$ as a stack of $k$ decomposable neural network blocks: 
\begin{equation}
\phi\coloneqq\phi^{\pi_0 \rightarrow \pi_k}=\phi^{\pi_0\rightarrow\pi_1}\circ \phi^{\pi_1\rightarrow\pi_2}\circ\cdots\circ\phi^{\pi_{k-1}\rightarrow\pi_k},
\end{equation}
where $\circ$ denotes the function composition and $\phi^{\pi_{j}\rightarrow\pi_{j+1}}$ indicates a block that recovers $\pi_{j+1}$-bit weights from $\pi_{j}$-bit weights. That is, we design the set of bitwidths for dequantizer blocks $\Pi$ to include the client bitwidths so that the dequantizer can directly reconstruct the desired bitwidths from the received low-bit weights. We implement each block using a dimensionality-preserving function $h(\cdot;\bsy\theta)$ with a residual connection to its input. 
When the model $(\bm{w}, \pi_{j>0})$ arrives from a local client, the server quantizes $\bm{w}$ into $j$ lower-bitwidth weights $\mathcal{Q}_{\bm{w}}=\{\bm{q}_{\pi_0},...,\bm{q}_{\pi_{j-1}}\}$. 
Given $\pi_{j<k}$-bit quantized weights $\bm{q}_{\pi_j}$, the block operation is formulated as follows:
\begin{equation}\label{eq:block}
\begin{split}
\widehat{\bm{q}}_{\pi_{j+1}}=\phi^{\pi_{j}}({\bm{q}}_{\pi_j};\bsy\theta_j)={\bm{q}}_{\pi_j} + h({\bm{q}}_{\pi_j};\bsy\theta_j).
\end{split}
\end{equation}
%When a model $(\bm{w}, \pi_{j>0})$ arrives from a local client, a server quantizes $\bm{w}$ into its lower-bitwidth weights $\mathcal{Q}_{\bm{w}}=\{\bm{q}_{\pi_0},...,\bm{q}_{\pi_{j-1}}\}$. 
The central server constructs a weight dataset out of received local model weights, by segmenting them into uniformly-sized blocks (e.g., $64\times24\times24$). Thus, our dequantizer can utilize differently-shaped weights from different layers or across heterogeneous neural architectures. We describe details of the weights dataset construction process and the design of network block $\phi$ in \Cref{appendix:dequantizer}. For the training of the weight dequantizer, we introduce two different loss terms. The first term is the reconstruction loss, which is defined as the average difference of the blockwise weight reconstruction of quantized input weights ${\bm{q}}_{\pi_j}$ and its higher bitwidth ground truth weights $\bm{q}_{\pi_{j+1}}$:
\begin{equation}\label{eq:loss_recon}
\begin{split}
%\mathcal{L}_{recon}=\sum^{k-1}_{i=0}\big\|\bm{w}^{(i+1)}_Q-\phi^{i\rightarrow i+1}(\bm{w}^{(i)}_Q)\big\|^2_2
\mathcal{L}_{recon}=\sum^{k-1}_{j=0}\big\|\bm{q}_{\pi_{j+1}}-\phi^{\pi_j\rightarrow\pi_{j+1}}({\bm{q}}_{\pi_j};\bsy\theta_j)\big\|_1.
\end{split}
\end{equation}
%We use the $\ell_1$ norm for measuring the difference.
Minimizing the blockwise reconstruction error allows the model to progressively extrapolate missing links between the low-bitwidth weights and high-bitwidth weights and not to stray far from the intermediate reconstructions. We note that there is no strict rule for the choices of the target bitwidths of dequantizer blocks, but we design the dequantizer in which the bitwidth difference between consecutive blocks is maintained to a similar degree until reaching the target high-bitwidth, avoiding drastic increase in bitwidth for each reconstruction step.

The second term is a distillation loss, for minimizing the descrepancy in the predictions from the model with high-bitwidth ground truth weights, and the model with dequantized low-bitwidth weights. 
%Here, we introduce the synthetic sample generator or an unsupervised validation buffer for the prediction of the central server since the server cannot receive the private local data in the federated learning scenario.
%We use a tiny buffer $\mathcal{U}$ at the server for computing distillation, which is independent of the local data. For each arrived weights $\bm{w}$, we randomly sample a minibatch of samples $\bm{u}\sim\mathcal{U}$, and compute the distillation loss as follows:
To this end, the central server utilizes a tiny buffer $\mathcal{U}$ independent of the local data to compare the prediction between high-bit model and recovered model from the low-bit weights. 
Given the $\bm{w}$ and its quantized low-precision $\bm{q}_{\pi_0}=Q_{\pi_0}(\bm{w})$, we compute a distillation loss using a randomly sampled minibatch $\bm{u}\sim\mathcal{U}$ from the buffer as follows:
\begin{equation}\label{eq:loss_distill}
\begin{split}
%\mathcal{L}_{distill}=-\sum^{k-1}_{i=0}\text{Sim}\Big(f(\bm{u}; \bm{w}), f\big(\bm{u}; \phi^{i\rightarrow {k}}(\widehat{\bm{w}_i};\bsy\theta_{i:k})\big)\Big),
\mathcal{L}_{distill}=-\text{Sim}\Big(f(\bm{u}; \bm{w}), f\big(\bm{u}; \phi^{0\rightarrow {k}}(\bm{q}_{\pi_0};\Theta)\big)\Big),
\end{split}
\end{equation}
%where $\bsy\theta_{i:k}=\{\bsy\theta_i,\cdots,\bsy\theta_k\}$ and $\text{Sim}$ is available to any type of similarity metric that we use cosine similarity.
where $\text{Sim}(\cdot)$ is a similarity metric for the two output distributions (we use cosine similarity) and $\Theta=\{\bsy\theta_0,\cdots,\bsy\theta_k\}$. Note that $f(\bm{u};\bm{w})$ is a softmax class probability of a neural network parameterized by $\bm{w}$ given input $\bm{u}$. %We emphasize that the distillation loss directly considers the reconstructed model prediction with generated weights, avoiding the dequantizer to fall into local reconstruction of the model fragments as negligent in the full model structure.
This distllation loss allows ProWD to directly tackle the prediction task with the reconstructed weights, which helps it recover full-precision weights while considering the downstream task performance.
The final objective for training ProWD at the server is given as follows:
\begin{equation}\label{eq:loss_final}
\begin{split}
\mathcal{L}=\mathcal{L}_{recon} + \lambda \mathcal{L}_{distill},
\end{split}
\end{equation}
where $\lambda$ is a hyperparameter to balance the two loss terms. Note that the model is not very sensitive to the choice of $\lambda$ and thus we set $\lambda=1$ in all our experiments. The training steps of our dequantizer is described in \Cref{algo:algorithm2}.
%we set $\lambda=1$ for all experiments regardless of varying degrees in bitwidth heterogeneity, demonstrating the efficiency in terms of the tuning of our method.
\input{contents/tables/3_algorithm_2}

\subsection{Score-based Selective Weight Aggregation}\label{subsec:sgu}
%We further propose a develpoed weight aggregation method to tackle the deteriorating shift in distributions during BHFL. The server selectively utilizes the received weights to encourage informative knowledge aggregation among different bitwidth weights. 
As described in \Cref{fig:challenge}, using a naive aggregation technique such as simple averaging, may lead the model training to fall into a suboptimal local minimum. 
To prevent such distribution shifts of weights during BHFL, we assert that the low-bitwidth models should share similar gradient directions as the full-precision model, as inspired by the observation in~\citet{zhu_towards_2020} that there exists a strong correlation between the training stability and the deviation of the quantized model's gradient directions from the full-precision model, measured by the cosine similarity. 
Let us consider a simple two-client federated learning framework where a server communicates with a full-precision model parameterized with $\bm{w}_\text{High}$ and a low-bitwidth model parameterized with $\bm{w}_\text{Low}$. The goal of the low-bitwidth model then is to distill the knowledge of $\bm{w}_\text{High}$:
\begin{equation}\label{eq:teacher_fp}
\begin{split}
\bigg<\frac{\partial\ell(f(B;\bm{w}_\text{High}))}{\partial\bm{w}_\text{High}},\frac{\partial_Q\ell(f_Q(B;\bm{w}_\text{Low}))}{\partial_Q\bm{w}_\text{Low}}\bigg>\geq0,
\end{split}
\end{equation}
where $\ell$ is a task loss and $f(\cdot;\bm{w})$ is a neural network parameterized by $\bm{w}$. The subscript $Q$ denotes the quantized operations on the weights, activations, and gradients.

However, unlike the setting of \citet{zhu_towards_2020}, under BHFL scenarios, it is impossible to preserve full-precision knowledge on low-bitwidth local devices as they have no means to represent them. Thus, we impose a selective weight aggregation technique based on the relevancy among the weights from the local clients. 
%we perform elementwise weight selection to maximize the similarity between the averaged weight movements of low-bitwidth and high bithwidth models.
When a central server receives the local models from different bitwidth clients, we select sparse sub-weights from lower-bitwidth models that are compatible with the weights of the full-precision aggregated weights.

Let $\overline{\bm{w}}_\text{High}^{(r)}$ and $\overline{\bm{w}}_{\text{Low}}^{(r)}$ denote average high-bit weights and the low-bit weights that the server received at communication round $r$, respectively. When $\Delta\overline{\bm{w}}_\text{High}=\overline{\bm{w}}^{(r)}_\text{High}-\overline{\bm{w}}_\text{High}^{(r-1)}$ and $\Delta\overline{\bm{w}}_\text{Low}=\overline{\bm{w}}^{(r)}_\text{Low}-\overline{\bm{w}}_{Low}^{(r-1)}$, we encourage the high-bit and low-bit model weights to have similar update directions by disregarding a few outliers in the low-bit models. That is, Given a sparsity ratio $\tau$, we encourage a server to obtain the binary mask $c^*$ as follows:
\begin{equation}\label{eq:ssa2}
\begin{split}
\bm{c}^*=\underset{\bm{c}}{\text{argmax}}\frac{(\bm{c}\odot\Delta\overline{\bm{w}}_\text{Low})^\top\Delta\overline{\bm{w}}_\text{High}}{\|\bm{c}\odot\Delta\overline{\bm{w}}_\text{Low}\|\|\Delta\overline{\bm{w}}_\text{High}\|},~\text{s.t.}~|\bm{c}^*|\leq\tau.\\
%~~~\text{or}~~\underset{\bm{c}}{\text{argmax}}\frac{(\bm{c}\odot\Delta\overline{\bm{w}}_{Q})^\top\cdot(\bm{c}\odot\Delta\overline{\bm{w}}_{F})}{\|\bm{c}\odot\Delta\overline{\bm{w}}_{Q}\|\cdot\|\bm{c}\odot\Delta\overline{\bm{w}}_{F}\|},\\
\end{split}
\end{equation}
For BHFL scenarios with multiple bitwidths, we split high-/low-bitwidths weights based on the mean bit-width of given a bitwidth set. We formulate the equation as a simple optimization problem to obtain a desired binary mask $c^*$ to maximize the cosine similarity between sparsified averaged weight movement of low-bitwidth models and the averaged
movement of high-bitwidth model. The process is rapidly optimized within a few steps (e.g., $10$) and a marginal training time {($\sim10ms$ per client)}. To this end, we perform selective weight aggregation as follows:
\begin{equation}\label{eq:aggregation}
\begin{split}
%\bm{w}_G\leftarrow\bm{w}_G+\frac{1}{|\bm{b}|}\sum^{|\bm{b}|}_{s\in\bm{b}}\kappa_s\cdot\bm{c}_s\odot\Delta\overline{\bm{w}}_s,
\bm{w}_G\leftarrow\frac{1}{N}\sum^{N}_{n=1}&\bm{c}_n\odot\bm{w}_n, %\\
%&\text{where}~~
%    \bm{c}_n=\begin{cases}
%        \bm{c}^* &, \text{if} \\
%        \bf{1} &, \text{otherwise}.\\
%    \end{cases}
\end{split}
\end{equation}
where $\bm{c}_n=\bm{c}^*$, if $s_n$ is one of low-bitwidth specifications, otherwise, $\bm{c}_n$ is a all-one tensor with the same shape as $\bm{w}_n$.
The overall procedure of our BHFL framework is described in \Cref{algo:algorithm1}.
%Note that the training of the dequantizer can be done in parallel to the weights reconstruction for aggregation during inference, preventing additional training time bottleneck.
Note that the dequantizer can be updated anytime during the FL process in a concurrent manner, because it does not require the latest client model weights for training. Thus, there is no training time bottleneck introduced by the training of our dequantizer.

\input{contents/tables/3_algorithm_1}

%% file: contents/tables/3_algorithm_2.tex
\begin{minipage}[t!]{\linewidth}
    %\vspace{-0.1in}
    \begin{algorithm}[H]%[htbp]
    \small
    \caption{Training of progressive weight dequantizer}
	\label{algo:algorithm2}
	\begin{algorithmic}[1]
	    \INPUT \small{set of bitwidths $\Pi=\{\pi_0, \dots, \pi_k\}$, dequantizer $\phi_{\Theta}^{\pi_0\rightarrow\pi_k}$, input weights and corresponding bitwidth $(\bm{w}, \pi_j)$, neural network $f$, unsupervised buffer $\mathcal{U}$, balancing coefficient $\lambda$.}
        %\STATE $\bm{q}_{j<j_w}=\{Q_{\pi_j}(\bm{w})\}_{j<j_w}$ \COMMENT{Quantize into each of the lower bitwidths}
        \STATE $\bm{q}_{i<j}=\{Q_{\pi_i}(\bm{w})\}_{i=0}^{j-1}$ \COMMENT{Quantize into each of the lower bitwidths}
        \STATE Construct data loader $\mathcal{D}_{\bm{w}}$ using $\{\bm{q}_{i<j}, \bm{w}\}$
        %\FOR{iteration $e=1,2, ..., E$}
        \FOR{$\left(\bm{q}_0, \dots, \bm{q}_{j-1},~\bm{w}\right)\sim\mathcal{D}_{\bm{w}}$}
            \STATE Sample $\bm{u}\sim\mathcal{U}$ augmented with gaussian noise
            \STATE $\mathcal{L}_{recon}=\sum^{j-1}_{i=0}\big\|\bm{q}_{i+1}-\phi^{\pi_i\rightarrow \pi_{i+1}}({\bm{q}}_i;\bsy\theta_{i})\big\|_1$%\COMMENT{\Cref{eq:loss_recon}}
            \STATE $\mathcal{L}_{distill}=-\text{Sim}(f(\bm{u};\bm{w}), f(\bm{u};\phi_{\Theta}(\bm{q}_0)))$%\COMMENT{\Cref{eq:loss_distill}}
            \STATE $\mathcal{L}=\mathcal{L}_{recon} + \lambda \mathcal{L}_{distill}$%\COMMENT{\Cref{eq:loss_final}}
            \STATE Update weight dequantizer $\phi_{\Theta}$ to minimize $\mathcal{L}$
        \ENDFOR
    \end{algorithmic}
	\end{algorithm}
	%\vspace{-0.25in}
\end{minipage}

%% file: contents/tables/3_algorithm_1.tex
\begin{minipage}[t!]{\linewidth}
% \vspace{-0.1in}
\begin{algorithm}[H]
    \small
    \caption{ProWD framework for BHFL}
	\label{algo:algorithm1}
	\begin{algorithmic}[1]
	    \INPUT clients $\mathcal{C}\leftarrow\{\bm{w}_n,s_n\}_{n=1}^N$, $s_j\in\Pi=\{\pi_0, \dots, \pi_k\},~\forall j$, weight dequantizer $\phi_{\Theta}^{\pi_0\rightarrow\pi_k}$, global weights $\bm{w}_G$. \\ 
	    \FOR{each round $r=1,2,..., R$}
            \STATE Sample $\mathcal{C}^{(r)}\subseteq\mathcal{C}$ where $|\mathcal{C}^{(r)}|=M$
            \STATE Distribute federated weights $\bm{w}_G$ to clients $\mathcal{C}^{(r)}$
            \FOR{each client $(\bm{w}_n,s_n)\in\mathcal{C}^{(r)}$ \textbf{in parallel}}
                \STATE $\bm{w}_n\leftarrow Q_{s_n}(\bm{w}_G)$ \COMMENT{send $s_n$-bit quantized weights}
                \STATE $\bm{w}_n\leftarrow$ LocalUpdate($\bm{w}_n,s_n$)
                \STATE Broadcast $(\bm{w}_n, s_n)$ to the central server
            \ENDFOR
            \STATE ${\bm{w}}_n\leftarrow \phi_{\Theta}(\bm{w}_n)$, \textbf{if} $s_n\neq\pi_k$ \COMMENT{Dequantize weights}
            \STATE Obtain binary masks $\bm{c}^*$ using \Cref{eq:ssa2}
            \STATE $\bm{w}_G\leftarrow\frac{1}{M}\sum^{M}_{n=1}\bm{c}_n\odot\bm{w}_n$ \COMMENT{\Cref{eq:aggregation}}
        \ENDFOR
	\end{algorithmic}
	\end{algorithm}
\end{minipage}\\

%% file: contents/4_experiments.tex
\input{contents/tables/5_cifar_table}

\input{contents/figures/5_main_plot}

\section{Experiments}
% We validate our method against the relevant FL methods under several BHFL scenarios, with varying bitwidth configurations of participating clients. We use the widely used benchmark dataset for federated learning methods, \textsc{CIFAR-10}, following the IID experimental settings of the existing works~\cite{reisizadeh_fedpaq_2020,haddadpour_federated_2020}. We use a modified version of VGG-7 network~\cite{simonyan_very_2015} for all our experiments.

\paragraph{Datasets} 
We validate our method against the relevant FL methods under several BHFL scenarios, with varying bitwidth configurations of participating clients. We use the widely used benchmark dataset for federated learning methods, \textsc{CIFAR-10} to validate our method following the IID experimental settings of the existing works~\cite{reisizadeh_fedpaq_2020,haddadpour_federated_2020}. \textsc{CIFAR-10} is a image classification dataset that consists of 10 object classes each of which has 5,000 training instances and 1,000 test instances. For FL purposes, we uniformly split the training instances per class by the number of clients participating in the federated learning system. We use a modified version of VGG-7 network~\cite{simonyan_very_2015} for all our experiments.

\paragraph{Data augmentation} We perform the standard random crop with 4-pixel padding followed by a random horizontal flip and a random rotation of maximum $15^\circ$ for all clients.

\paragraph{Baselines} We compare our method \textit{ProWD} against following FL baselines:
\eat{standard FL methods:~\textbf{1) FedAvg}~\citep{McMahan2017} and \textbf{2) FedProx}~\citep{li2018federated},
FL methods with Quantized Parameter Communication (QPC):~\textbf{3) FedPAQ}~\citep{reisizadeh_fedpaq_2020}, \textbf{4) FedCOM}, and \textbf{5) FedCOMGATE}~\citep{haddadpour_federated_2020}, naive solutions for BHFL scenarios: \textbf{6) FedGroupedAvg} and \textbf{7) FedGroupedAvg-Asymmetric}.}
% \vspace{-0.3in}
% \begin{itemize}[leftmargin=0.in]
% \setlength\itemsep{0.3em}
\textbf{FedAvg}~\citep{McMahan2017}: A popular federated learning method that performs simple averaging of the local models at the server at each round.
\textbf{FedProx}~\citep{li2018federated}: A federated learning method that aims to deal with device heterogeneity, with additional $\ell_2$ distance regularization terms during the update of local models to prevent the model divergence.
\textbf{FedPAQ}~\citep{reisizadeh_fedpaq_2020}: A quantized parameter communication (QPC)-based FL method, which at each round quantizes the difference between the current local weights and the last aggregated weights at each client, then sends it to the server for aggregation.
\textbf{FedCOM}~\citep{haddadpour_federated_2020}: An extension of FedPAQ with a learnable global learning rate, which achieves faster convergence in data-homogeneous settings.
\textbf{FedCOMGATE}~\citep{haddadpour_federated_2020}: An extension of FedCOM to data-heterogeneous settings, which uses a correction vector that constrains the local models to evolve in a similar direction.
\textbf{FedGroupedAvg}: A variant of FedAvg in which the server separately aggregates weights only from the same bitwidth clients, and redistributes them to corresponding local clients.
\textbf{FedGroupedAvg-Asymmetric}: A modified version of FedGroupedAvg that sends clients only the aggregated weights of other clients that has the same bitwidth or higher.
% \end{itemize}

Due to the page limit, we provide the details of hyperparameters for baselines and ours in \Cref{appendix:exp_setup}.

\subsection{Quantitative Evaluation}
We validate our methods under multiple BHFL scenarios with heterogeneous proportions of bitwidths among the clients. We first report the experimental results with 50$\%$ of Int8 and 50$\%$ Float32 clients (Left) and 80$\%$ of Int8 and 20$\%$ Float32 clients (Right) in \Cref{tab:main}. %Local training, where each client trains independently on its local task, to understand the effect of knowledge sharing during BHFL. 
Int8 models in FedAvg obtain superior performance to local training, where each client trains independently on its local task due to positive knowledge transfer from the full-precision weights. This is also evident in the poor performance of Int8 clients in FedGroupedAvg which demonstrates that FL only with low-bitwidth clients is highly limited due to the lack of information the low-bit weights provide. 
QPC-based FL methods, FedPAQ, FedCOM, and FedCOMGATE, communicate the quantized form of accumulated gradients at each round, and the server adds the accumulated gradients to the global model at each round, before broadcasting it to the clients. Such gradient communication does not degrade the local task information during FL, which is helpful for Float32 models to mitigate the interference from the low-bit weights.
However, the low-bit clients (e.g., Int8) cannot directly amalgamate the expressive knowledge from the high-bitwidth weights, easily falling into suboptimal local minima.
FedGroupedAvg variants, while preserving the performance of the high-bit models, they often obtain inferior performance as they do not exploit the full knowledge provided by other models. 
On the other hand, our ProWD consistently outperforms all baselines  with varying compositions of bitwidths, yielding small performance gap between low- and high-bitwidths, which demonstrates the effectiveness of the proposed progressive weight dequantization scheme with selective weight aggregation. The convergence plot in \Cref{fig:main_plot} shows that our method rapidly converges to good performance while baselines converge to suboptimal local minima. 

We further demonstrate the versatility of the ProWD framework under a BHFL scenario with multiple bitwidths in \Cref{tab:multi-scaled-main}, using devices with diverse bitwidths. Note that we allow local clients to send the parameters with local bitwidth for this experiment, rather than ternarizing them for uplink communication. Int6 and Int8 are considered as low-bitwidths and Int12 and Int16 are considered as high-bitwidths. Our ProWD consistently outperforms baselines with any bitwidths while achieving a small accuracy gap between Int6 and Int16 clients. We expect that the reduced performance gap between ours and FedAvg is due the smaller disparity between weight distributions compared to those in \Cref{tab:main} (Int8$\leftrightarrow$Float32), since all clients perform low-bit operations with slightly different bitwidths, in which case FedAvg may suffer less from distributional shift at aggregation. To validate that, we further provide additional experiments with larger variance among the bitwidths of the participating devices in~\Cref{fig:multibits}, which shows a significant performance gain of our ProWD to FedAvg (Int6 clients acc: \textbf{5.9}\%$\bsy\uparrow$, average acc: \textbf{2.7}\%$\bsy\uparrow$, performance gap: \textbf{36}\%$\bsy\downarrow$).
\input{contents/tables/5_multiple_bits_table}
\input{contents/tables/5_ablation_study}

\paragraph{Ablation study}
Now we explicate the effect of each ingredient of our ProWD on the CIFAR-$10$ with an ablation study. We experiment on both the uniform ($50\%$ of Float32 and $50\%$ of Int8 clients) and low-bitwidth dominant scenario ($20\%$ of Float32 and $80\%$ of Int8 clients). As shown in \Cref{tab:ablation}, thanks to its ability to reconstruct full-precision weights from low-bitwidth weights, our weight dequantizer (Deq) improves the performance by $5.5\%$ and {$2.8\%$} over that of FedAvg, respectively for each scenario. Also, our selective weight aggregation (SWA) minimizes the discrepancy across the model updates from different bitwidth clients, obtaining the significant performance gain of $9.2\%$ and {$2.6\%$} for each scenario. This experimental result confirms the efficacy of both components, progressive weight decomposition, and selective weight aggregation.

\subsection{Qualitative Analysis}
\paragraph{The effect of progressive weight dequantization}
To further analyze the role of the progressive weight dequantizer in our method, we visualize stepwise distributions of reconstructed weights using our dequantizer. For ease of interpretation, we use the initial input as the quantized weight from the last convolution layer in the Int8 client for training on CIFAR-10, where the result is illustrated in \Cref{fig:deq_anal}. As low-bitwidth clients transmit ternarized model weights to the central server, the distribution of input weights to the dequantizer is visualized in three peaks. We also visualize the distribution from reconstructed weights after forwarding initial weights up to the first, second, and last block in our dequantizer, colored by \emph{orange}, \emph{pink}, and \emph{navy}, respectively. As we expected, our dequantizer progressively recovers ternary weights towards the high-bitwidths by forwarding them through sequential dequantization blocks.

% \vspace{-0.1in}
\paragraph{Weight distance between local clients}
Next, we visualize the distance of model weights between clients in \Cref{fig:distance} to dissect the difference of learned representations during BHFL. We use the cosine distance, computed by $1 - \text{cos\_sim}(\bm{w}_n, \bm{w}_m)$ between $n$- and $m$-th local clients. Weights obtained using FedAvg have smaller distances between Int8 and Float32 models than the distances among the Int8 models, caused by the distributional shift of the high-bitwidth model weights towards the distribution of low-bitwidth model weights (Please see \Cref{fig:challenge}). 

Clients in FedPAQ communicate accumulated local gradients while keeping their local weights, alleviating the detrimental distributional shift of weights to some degree, resulting in a bigger similarity among the same bitwidth clients than in bit-heterogeneous cases.
Int8 client weights in FedGroupedAvg stay close to each other while straying far from the Float32 client weights, which is expected as it only allows communication among clients with the same bitwidth.
Interestingly, ProWD keeps the clients' weights sufficiently different between each bitwidthobtains sufficiently low proximity across bit-heterogeneous clients, successfully preventing the distributional shift in high-precision weights due to weight averaging. 
Our proposed method also allows high transferability of the learned knowledge across clients that shows better adaptation on the local tasks over QPC-based methods (Please see \Cref{tab:main} and \Cref{fig:main_plot}). We provide more similarity analysis of the local models for other baselines and our models in \Cref{appendix:add_experiments}.
\input{contents/figures/5_deq_anal}
\input{contents/figures/5_distance_matrix}

%% file: contents/tables/5_cifar_table.tex
\begin{table*}[t]
% \vspace{-0.05in}
\caption{\small Average accuracy at each bitwidth and average accueacy across all clients on CIFAR-10 dataset. We set participating clients with 50\% of Int8 and 50\% of Float32 models (Left), and 50\% of Int8 and 50\% of Float32 models (Right). All of the results are measured by computing the 95\% confidence interval over three independent runs.}
\label{tab:main}
\vspace{-0.05in}
\centering
\small
\setlength{\tabcolsep}{3pt} % 
\resizebox{0.95\textwidth}{!}{
\begin{tabular}{l@{\hspace{6pt}}c c c  a c c c c a}
%\begin{tabular}{llcccccc}
\toprule
{\textsc{Method}}&\multicolumn{4}{c}{\textbf{\textsc{CIFAR-10}, {Int8~(50\%)~-~{Float{32}~(50\%)}}}} &~& \multicolumn{4}{c}{\textbf{\textsc{\textbf{CIFAR-10}, {Int8~(80\%)}~-~{Float{32}~(20\%)}}}}\\
\midrule
%& & Accuracy ($F$) & Accuracy ($I$) & Averaged & Accuracy ($F$) & Accuracy ($I$) & Averaged\\
%& & $F32$ Accuracy & $I8$ Accuracy & Average & $F32$ Accuracy & $I8$ Accuracy & Average\\
& \textsc{int8 acc} & \textsc{float32 acc} & \textsc{gap} &\textsc{average} && \textsc{int8 acc} & \textsc{float32 acc} & \textsc{gap} & \textsc{average}\\
\midrule
%\parbox[t]{2mm}{\multirow{7}{*}{\rotatebox[origin=c]{90}{\small  $\bm{F{32}~(5)}~-~\bm{I8~(5)}$}}}
%\parbox[t]{2mm}{\multirow{9}{*}{\rotatebox[origin=c]{90}{\small  \textsc{{F{32}~(50\%)}~-~{Int8~(50\%)}}}}}
\textsc{Local Training} & 
{69.59}~\scriptsize($\pm$~0.34) &
{75.82}~\scriptsize($\pm$~0.41) &
\plhi{+6.23} & {72.71}~\scriptsize($\pm$~0.26) &&
{69.39}~\scriptsize($\pm$~0.34) &
{76.41}~\scriptsize($\pm$~0.52) &
\plhi{+7.02} & {70.79}~\scriptsize($\pm$~0.34)\\
\midrule
\textsc{FedAvg}~\citep{McMahan2017} & 
{76.88}~\scriptsize($\pm$~0.49) &
{76.23}~\scriptsize($\pm$~0.36) &
\mihi{-0.65} & {76.56}~\scriptsize($\pm$~0.36) &&
{77.43}~\scriptsize($\pm$~0.83) &
{74.64}~\scriptsize($\pm$~1.21) &
\mihi{-2.79} & {76.87}~\scriptsize($\pm$~0.87)\\
\textsc{FedProx}~\citep{li2018federated} & 
{71.16}~\scriptsize($\pm$~0.35) &
{69.28}~\scriptsize($\pm$~0.90) &
\mihi{-1.88} & {70.22}~\scriptsize($\pm$~0.55) &&
{69.60}~\scriptsize($\pm$~0.50) &
{66.28}~\scriptsize($\pm$~1.24) &
\mihi{-3.32} & {68.94}~\scriptsize($\pm$~0.57)\\
\textsc{FedPAQ}~\citep{reisizadeh_fedpaq_2020} & 
{78.01}~\scriptsize($\pm$~0.55) &
{84.93}~\scriptsize($\pm$~0.30) &
\plhi{+6.93} & {81.47}~\scriptsize($\pm$~0.29) &&
{76.61}~\scriptsize($\pm$~0.60) &
{82.27}~\scriptsize($\pm$~0.42) &
\plhi{+5.66} & {77.74}~\scriptsize($\pm$~0.49)\\
\textsc{FedCOM}~\citep{haddadpour_federated_2020} & 
{75.37}~\scriptsize($\pm$~0.52) &
{77.69}~\scriptsize($\pm$~0.45) &
\plhi{+2.32} & {76.53}~\scriptsize($\pm$~0.38) &&
{73.60}~\scriptsize($\pm$~1.08) &
{80.73}~\scriptsize($\pm$~0.73) &
\plhi{+7.12} & {75.03}~\scriptsize($\pm$~0.94)\\

\textsc{FedCOMGATE}~\citep{haddadpour_federated_2020} & 
{76.74}~\scriptsize($\pm$~0.59) &
{77.36}~\scriptsize($\pm$~0.55) &
\plhi{+0.62} & {77.05}~\scriptsize($\pm$~0.36) &&
{74.48}~\scriptsize($\pm$~0.63) &
{81.18}~\scriptsize($\pm$~0.56) &
\plhi{+6.70} & {75.82}~\scriptsize($\pm$~0.53)\\

\textsc{FedGroupedAvg} & 
{61.85}~\scriptsize($\pm$~0.78) &
{85.08}~\scriptsize($\pm$~0.28) &
\plhi{+23.23} & {73.46}~\scriptsize($\pm$~0.29)&&
{71.76}~\scriptsize($\pm$~0.68) &
{78.07}~\scriptsize($\pm$~0.58) &
\plhi{+6.31} & {73.02}~\scriptsize($\pm$~0.65) \\

\textsc{FedGroupedAvg-Asymmetric} &
{78.39}~\scriptsize($\pm$~0.54) &
{84.97}~\scriptsize($\pm$~0.27) &
\plhi{+6.57} & {81.68}~\scriptsize($\pm$~0.26)&&
{70.14}~\scriptsize($\pm$~0.36) &
{78.70}~\scriptsize($\pm$~0.48) &
\plhi{+8.56} & {71.85}~\scriptsize($\pm$~0.36) \\

\midrule
%& \textsc{HIVE w/o dequantizer (Ours)} & 
%{79.79} \scriptsize($\pm$ 0.98) & \textbf{85.94} \scriptsize($\pm$ 0.92) & {82.84} %\scriptsize($\pm$ 0.55) &&
%{*} \scriptsize($\pm$ *) & {*} \scriptsize($\pm$ *) & {*} \scriptsize($\pm$ *) \\
\textsc{ProWD (Ours)} & 
\textbf{82.87}~\scriptsize($\pm$~0.37) &
\textbf{85.99}~\scriptsize($\pm$~0.20) &
\plhi{+3.12} & \textbf{84.43}~\scriptsize($\pm$~0.22)&&
\textbf{79.23} \scriptsize($\pm$~0.25) &
{81.26} \scriptsize($\pm$~0.45) &
\plhi{+2.03}& \textbf{79.63} \scriptsize($\pm$~0.23)\\
\bottomrule
\end{tabular}}
\end{table*}

%% file: contents/figures/5_main_plot.tex
\begin{figure*}[t!]
    % \vspace{-0.15in}
    \centering
    \resizebox{\linewidth}{!}{%
    \begin{tabular}{cc}
    \includegraphics[width=8cm]{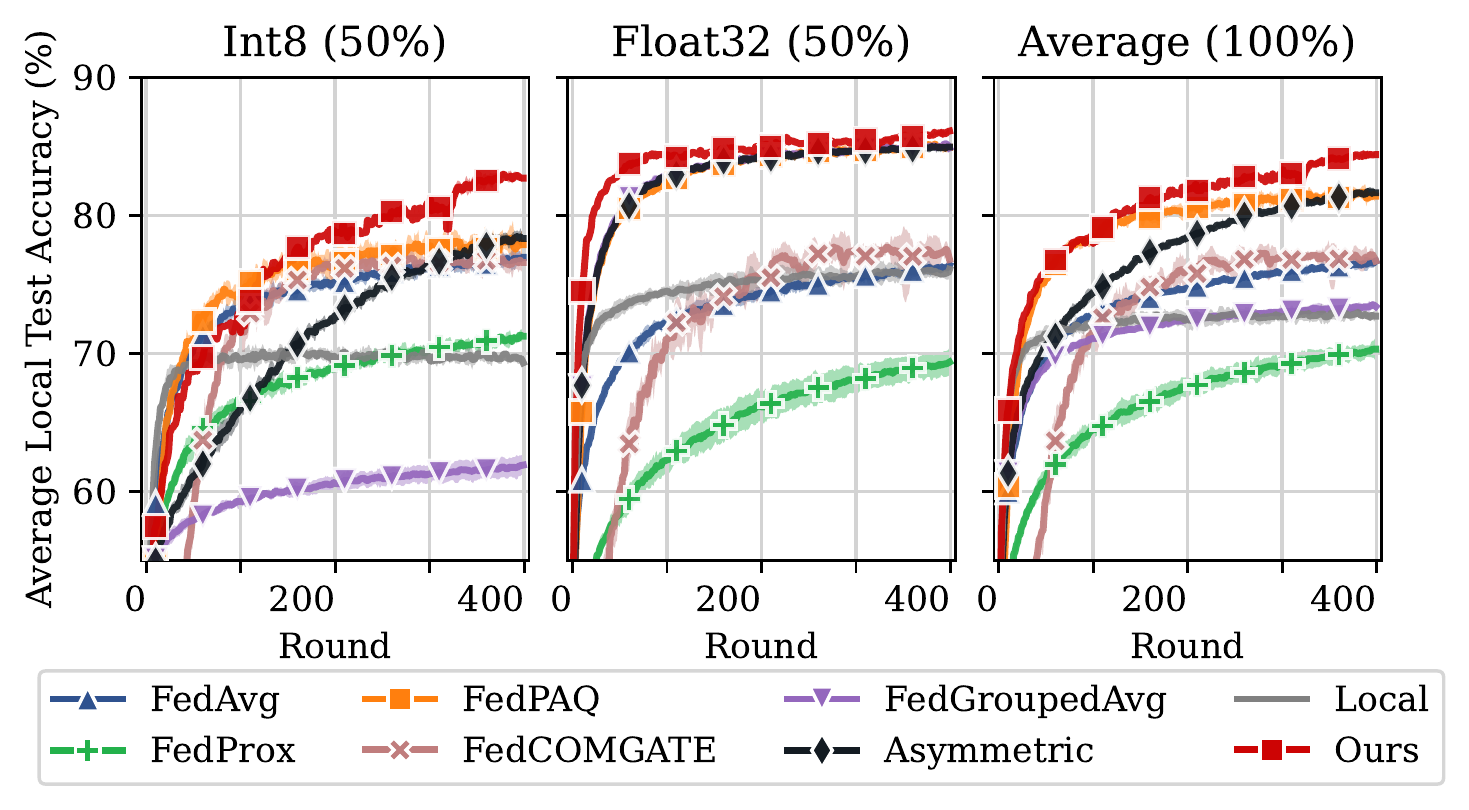}&\hspace{-0.1in}
    \includegraphics[width=8cm]{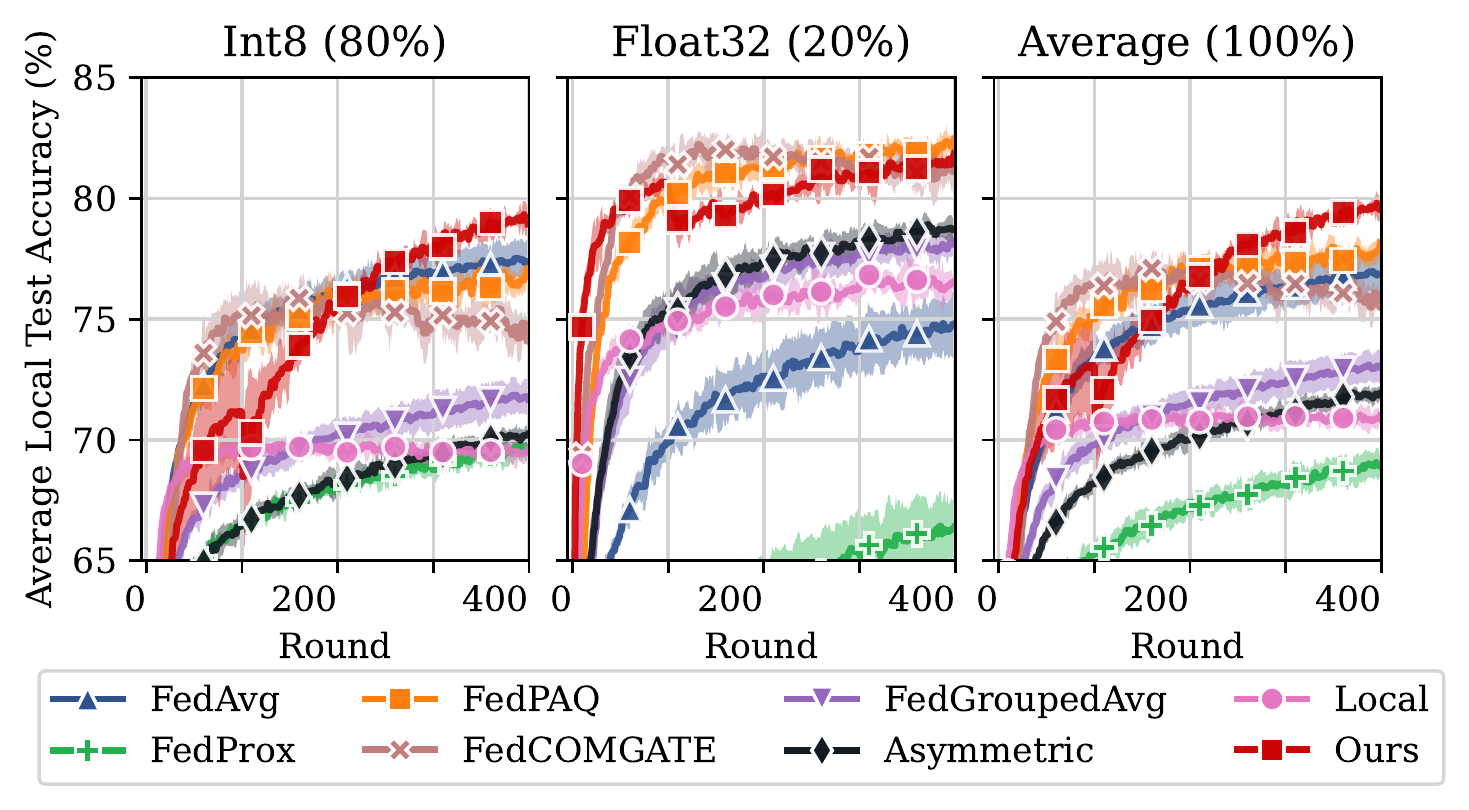} \vspace{-0.05in}\\
    %\vspace{-0.1in}
    %(a) CIFAR-10 & (b) FEMNIST \\
    \end{tabular}}
    \vspace{-0.1in}
    \caption{\footnotesize  Visualization of average test accuracy on CIFAR-10 with 10 clients where the bitwidth of the clients is composed of \textbf{(a) 50\% of Int8 and 50\% of Float32} and \textbf{(b)  80\% of Int8 and 20\% of Float32} during BHFL. We average the results over three independent runs.}
    %\caption{\footnotesize  Visualization of average test accuracy on \TBD{\textbf{(a) CIFAR-10 with 10 clients} and \textbf{(b) FEMNIST datasets with 50 clients} during BHFL, where the bitwidth of clients is composed of 50\% of Int8 and 50\% of Float32.} We average the results over three independent runs.}
    \label{fig:main_plot}
    \vspace{-0.1in}
\end{figure*}

%% file: contents/tables/5_multiple_bits_table.tex
\begin{table}[t]
\begin{minipage}[b]{\linewidth}
\centering
\small
%\caption{\small Multiple bit-width clients $F32 (2), I16 (2), I8 (3),$ and $I6 (3)$, $10$ clients in total.}
\caption{\small Average accuracy at each bitwidth, and across all clients on CIFAR-10. We set clients with 30\% of Int6, 30\% of Int8, 20\% of Int12, and 20\% of Int16 models. All the results are measured by computing and standard deviation over three independent runs.}
\label{tab:multi-scaled-main}
% \vspace{-0.1in}
\setlength{\tabcolsep}{3pt} % 
\resizebox{\textwidth}{!}{

\begin{tabular}{l@{\hspace{6pt}}ccccca}
\toprule
%{\textsc{Method}}&\multicolumn{5}{c}{\textbf{\textsc{CIFAR-10}}}\\
%\midrule
%& \textsc{int6 acc} & \textsc{int8 acc} & \textsc{int16 acc} & \textsc{float32 acc} & \textsc{Average}\\
{\textsc{Method}}& \textbf{\textsc{int6}} & \textbf{\textsc{int8}} & \textbf{\textsc{int12}} & \textbf{\textsc{int16}} & \textbf{\textsc{gap}}&\textbf{\textsc{AvgAcc}}\\
\midrule
\textsc{Local Training} & 
{69.51}&%~\scriptsize($\pm$~0.00) &        
{69.25}&%~\scriptsize($\pm$~0.00) &        
{68.96}&%~\scriptsize($\pm$~0.00) &        
{70.40}&%~\scriptsize($\pm$~0.00) &    
\plhi{+0.89}&
{69.50}\\%~\scriptsize($\pm$~0.00) \\
\midrule

\textsc{FedAvg}%~\citep{McMahan2017} 
& 
{80.17}&%~\scriptsize($\pm$~2.31) &
{85.71}&%~\scriptsize($\pm$~0.63) &
\textbf{86.49}&%~\scriptsize($\pm$~0.93) &
{87.02}&%~\scriptsize($\pm$~1.02) &
\plhi{+6.85}&
{84.47}\\%~\scriptsize($\pm$~0.76)\\

\eat{\textsc{FedProx}%~\citep{li2018federated} 
& 
{34.77}&%~\scriptsize($\pm$~0.00) &        
{66.12}&%~\scriptsize($\pm$~0.00) &        
{63.01}&%~\scriptsize($\pm$~0.00) &        
{61.41}&%~\scriptsize($\pm$~0.00) &        
\plhi{+26.64}&
{55.15}\\%~\scriptsize($\pm$~0.00) \\
}
\textsc{FedPAQ}%~\citep{reisizadeh_fedpaq_2020} 
& 
{80.78}&%~\scriptsize($\pm$~0.00) &        
{85.30}&%~\scriptsize($\pm$~0.00) &        
{85.98}&%~\scriptsize($\pm$~0.00) &        
{86.93}&%~\scriptsize($\pm$~0.00) &        
\plhi{+6.15}&
{84.40}\\%~\scriptsize($\pm$~0.00) \\

\textsc{FedCom}%~\citep{haddadpour_federated_2020} 
& 
{39.47}&%~\scriptsize($\pm$~0.00) &        
{55.14}&%~\scriptsize($\pm$~0.00) &        
{53.07}&%~\scriptsize($\pm$~0.00) &        
{53.26}&%~\scriptsize($\pm$~0.00) &        
\plhi{+13.79}&
{49.65}\\%~\scriptsize($\pm$~0.00) \\

\textsc{FedComGate}%~\citep{haddadpour_federated_2020} 
& 
{56.74}&%~\scriptsize($\pm$~4.07) &
{68.43}&%~\scriptsize($\pm$~2.77) &
{67.63}&%~\scriptsize($\pm$~2.62) &
{68.14}&%~\scriptsize($\pm$~2.37) &
\plhi{+11.69} &
{64.70}\\%~\scriptsize($\pm$~2.84)\\%~\scriptsize($\pm$~0.00) \\

\textsc{FedGroupedAvg} & 
{79.27}&%~\scriptsize($\pm$~2.25) &        
{80.62}&%~\scriptsize($\pm$~0.22) &        
{76.94}&%~\scriptsize($\pm$~0.20) &        
{77.39}&%~\scriptsize($\pm$~1.08) &        
\mihi{-1.88}&
{78.83}\\%~\scriptsize($\pm$~0.63) \\

\textsc{FedGroup-Asym} & 
{80.25}&%~\scriptsize($\pm$~0.00)
{79.74}&%~\scriptsize($\pm$~0.00) &
{79.18}&%~\scriptsize($\pm$~0.00) &
{77.73}&%~\scriptsize($\pm$~0.00) &
\mihi{-2.52}&
{79.38}%~\scriptsize($\pm$~0.00) &
\\%~\scriptsize($\pm$~0.00) \\

\midrule
\textsc{ProWD (Ours)} & 
\textbf{82.89}&%~\scriptsize($\pm$~0.00) &
\textbf{86.11}&%~\scriptsize($\pm$~0.00) &
\textbf{86.47}&%~\scriptsize($\pm$~0.00) &
\textbf{87.51}&%~\scriptsize($\pm$~0.00) &
\plhi{+4.62}&
\textbf{85.50}\\%~\scriptsize($\pm$~0.00) &
\bottomrule
\end{tabular}}
\end{minipage}
% \vspace{-0.05in}
\end{table}

\begin{figure}[t]
    \vspace{-.1in}
    \centering
    \includegraphics[width=\linewidth]{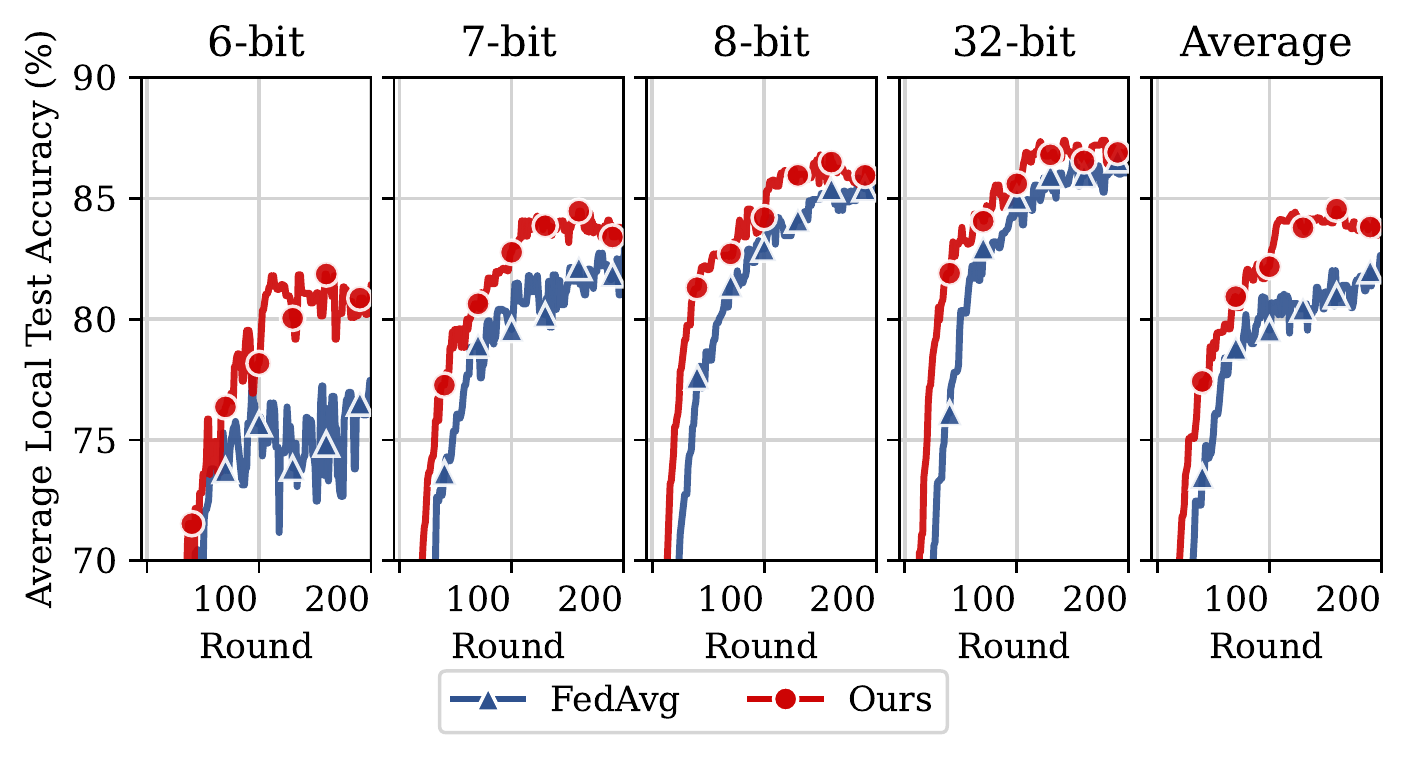}
    \vspace{-.3in}
    \caption{BHFL with Int6, Int7, Int8, and Float32 clients.}
    \label{fig:multibits}
    \vspace{-.1in}
\end{figure}

%% file: contents/tables/5_ablation_study.tex
\begin{table}[t]
% \vspace{-0.05in}
\caption{\small Ablation study for ProWD. \textsc{Deq} and \textsc{SWA} refer the progressive weight dequantizer and selevtive weight aggregation, respectively. We report the results over three independent runs.}
\label{tab:ablation}
% \vspace{-0.1in}
\centering
\small
\setlength{\tabcolsep}{3pt} % 
\resizebox{\linewidth}{!}{
\begin{tabular}{lcc@{\hspace{6pt}}c c a}
\toprule
{\textsc{Method}}&&\multicolumn{4}{c}{\textsc{\textbf{CIFAR-10}, int8 (50\%) - Float32 (50\%)}}\\
\midrule
& \textsc{Deq} & \textsc{SWA} & \textsc{int8 acc} & \textsc{float32 acc} & \textsc{average}\\
\midrule
\textsc{FedAvg} & $-$ & $-$ &
{76.55} \scriptsize($\pm$ 0.44) &
{75.71} \scriptsize($\pm$ 0.32) &
{75.83} \scriptsize($\pm$ 0.41) \\

\textsc{+Deq} & $\checkmark$ & $-$ &
{79.60} \scriptsize($\pm$ 0.39) &
{80.38} \scriptsize($\pm$ 0.44) &
{79.99} \scriptsize($\pm$ 0.26) \\

\textsc{+SWA} & $-$ & $\checkmark$ &
{79.79} \scriptsize($\pm$ 0.98) &
\textbf{85.94} \scriptsize($\pm$ 0.92) &
{82.84} \scriptsize($\pm$ 0.55) \\

\textsc{Ours} & $\checkmark$ & $\checkmark$ &
\textbf{82.87} \scriptsize($\pm$ 0.37) &
\textbf{85.99} \scriptsize($\pm$ 0.20) &
\textbf{84.43} \scriptsize($\pm$ 0.22) \\
\bottomrule
\toprule
{\textsc{Method}}&&\multicolumn{4}{c}{\textsc{\textbf{CIFAR-10}, int8 (80\%) - Float32 (20\%)}}\\
\midrule
& \textsc{Deq} & \textsc{SWA} & \textsc{int8 acc} & \textsc{float32 acc} & \textsc{average}\\
\midrule
\textsc{FedAvg} & $-$ & $-$ &
{77.43} \scriptsize($\pm$ 0.83) &
{74.64} \scriptsize($\pm$ 1.21) &
{76.87} \scriptsize($\pm$ 0.87) \\

\textsc{+Deq} & $\checkmark$ & $-$ &
{78.52} \scriptsize($\pm$ 0.50) &
{76.08} \scriptsize($\pm$ 0.44) &
{79.03} \scriptsize($\pm$ 0.31) \\

\textsc{+SWA} & $-$ & $\checkmark$ &
{78.42} \scriptsize($\pm$ 0.47) &
{80.82} \scriptsize($\pm$ 0.31) &
{78.90} \scriptsize($\pm$ 0.37) \\

\textsc{Ours} & $\checkmark$ & $\checkmark$ &
\textbf{79.23} \scriptsize($\pm$~0.25) &
\textbf{81.26} \scriptsize($\pm$~0.45) &
\textbf{79.63} \scriptsize($\pm$~0.23) \\
\bottomrule
\end{tabular}}
% \vspace{-0.1in}
\end{table}

%% file: contents/figures/5_deq_anal.tex
\begin{figure}[t!]
    % \vspace{-0.1in}
    \centering
    \small
    \begin{tabular}{c}
    \includegraphics[width=7cm]{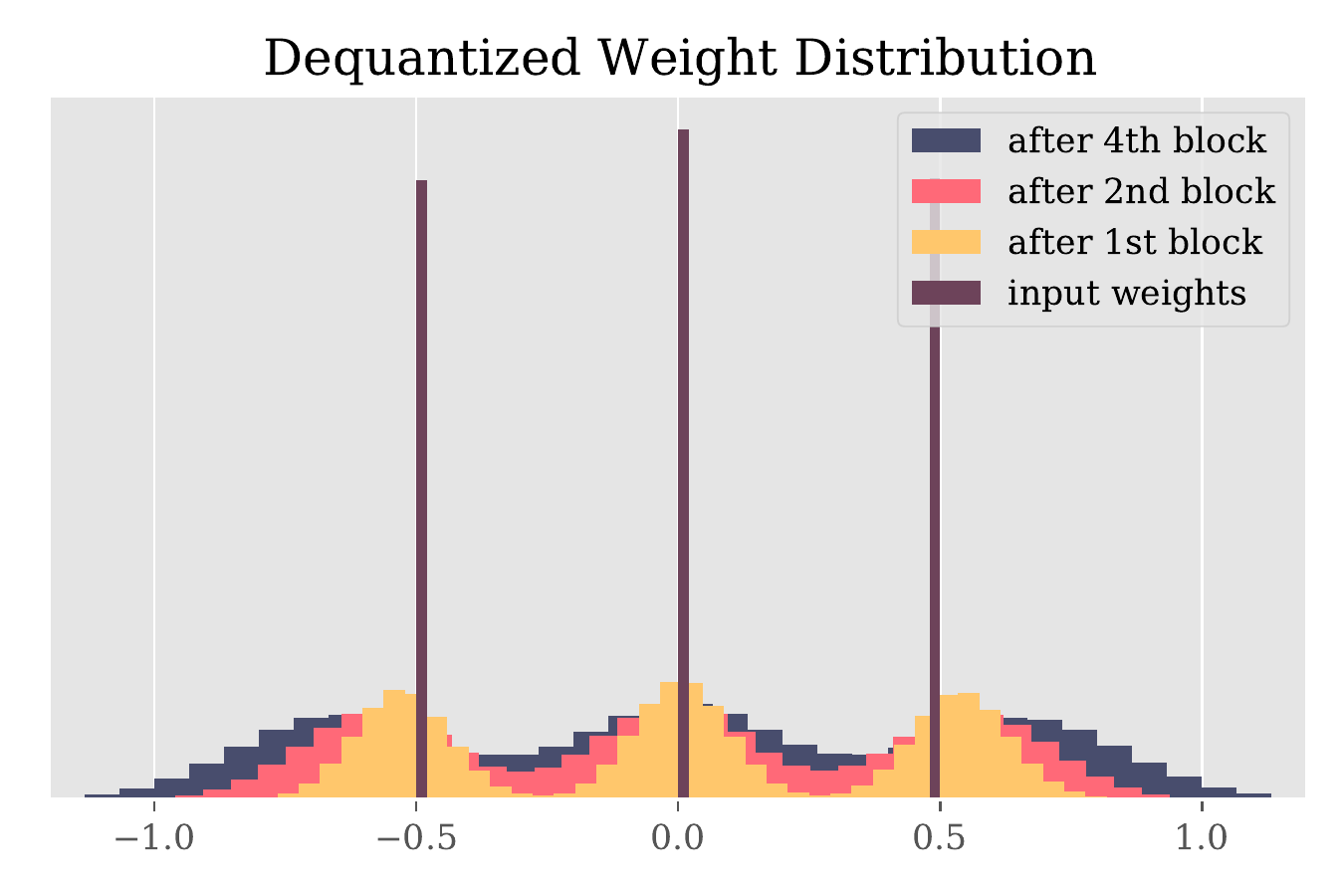}\\
    \end{tabular}
    \vspace{-0.2in}
    \caption{\footnotesize \textbf{The Weight Distribution after Progressive Dequantization.} Visualization of the distribution after the reconstruction of low-bitwidth model weights. %\TBD{'Initial' denotes ternary weights (input) and $0\rightarrow\#$ are recovered weights as the output of $\#^{th}$ blocks in the dequantizer.} 
    We visualize the last Convolution layer weights in neural network, trained on CIFAR-10.}
    \label{fig:deq_anal}
    \vspace{-0.05in}
\end{figure}

%% file: contents/figures/5_distance_matrix.tex
\begin{figure}
% \vspace{-0.05in}
\renewcommand{\tabcolsep}{1pt}
\begin{minipage}[c]{0.99\linewidth}
\begin{tabular}{cc}
\begin{overpic}[width=0.9\textwidth]{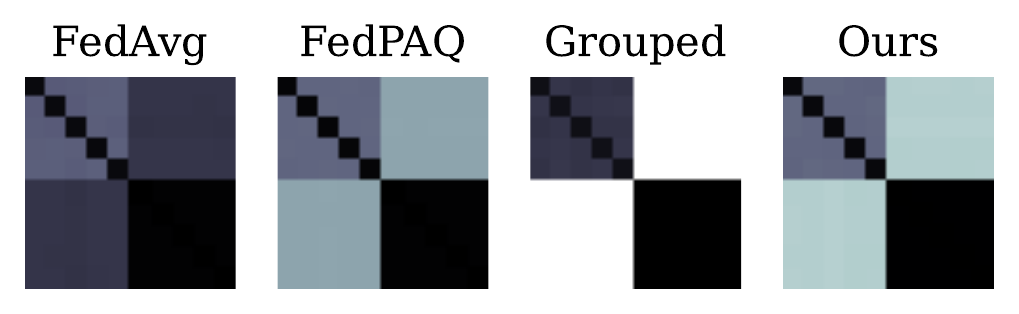}
\end{overpic} &
\begin{overpic}[width=0.1\textwidth]{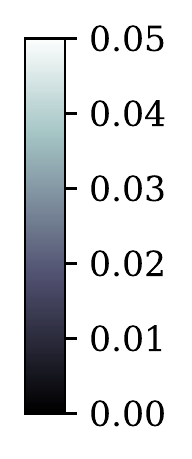}
\end{overpic}
\end{tabular}
\vspace{-0.15in}
\captionof{figure}{\small \textbf{Cosine distance matrix between the weights of each client.} Elements in a row and column describe the index of local clients. We set the first five clients to Int8 (Top left) and the other five to Float32 (Bottom right) for training on CIFAR-10. Darker colors indicate a bigger similarity.}
\label{fig:distance}
% \vspace{-0.1in}
\end{minipage}
\end{figure}

%% file: contents/5_conclusion.tex
\vspace{-0.1in}
\section{Conclusion}
We proposed a novel yet practical heterogeneous federated learning scenario where the participating devices have different bitwidth specifications, in which case the model aggregation could have a highly detrimental effect. We further identify the two main causes of the performance degeneration, which are the suboptimal loss convergence due to distributional shift from aggregation, and the limited expressive power of the low-bitwidth weights. To tackle these challenges, we proposed a novel framework for bit-heterogeneous FL, based on progressive dequantization of the weights and selective weight aggregation. The progressive dequantizer at the server receives weights from low-bitwidth clients and recovers them into higher bitwidth weights by forwarding them through a sequence of dequantizer blocks. Further, selective weight aggregation determines which low-bit weight elements are compatible with higher-bit ones. Empirical evaluations of our framework across two BHFL scenarios with varying degrees of bitwidth-heterogeneity on the benchmark dataset demonstrate the effectiveness of our framework, which largely outperforms relevant FL baselines.

\section{Acknowledgement}
{This work is supported by Samsung Advanced Institute of Technology, Institute of Information \& communications Technology Planning \& Evaluation (IITP) grant funded by the Korea government(MSIT)  (No.2019-0-00075, Artificial Intelligence Graduate School Program(KAIST)), and the Engineering Research Center Program through the National Research Foundation of Korea (NRF) funded by the Korean Government MSIT (NRF-2018R1A5A1059921).}

% the Engineering Research Center Program through the National Research Foundation of Korea (NRF) funded by the Korean Government MSIT (NRF-2018R1A5A1059921) and Institute of Information & communications Technology Planning & Evaluation (IITP) grant funded by the Korea government(MSIT) (No.2019-0-00075, Artificial Intelligence Graduate School Program(KAIST)).

%% file: contents/A_appendix.tex
\clearpage
\onecolumn
\appendix

\section{Hyperparameters}
\label{appendix:exp_setup}

At each round, we train each client for 200 local steps. The Float32 network is trained with SGD with learning rate 0.1, and momentum value 0.9. Additionally, the gradient $\ell_2$ norm is clipped to 2.0. The quantized fixed-point (Int2, Int4, Int8, Int12, and Int16) networks are trained with the low-bitwidth training method detailed in \Cref{appendix:low_bit_traing}, with $\eta=8.0$. For FedProx, we explored several l2 coefficients (0.01, 0.1, and 1), and found that 0.01 had the best final test accuracy. For FedCOM and FedCOMGATE, we use the global step size $\gamma = 10$, since \citet{haddadpour_federated_2020} only specifies that $\gamma$ should be greater than or equal to the number of clients participating in the FL process. For FedGroupedAvg and FedGroupedAvg-Asymmetric, for fair comparison, we scale down the local learning rates by the number of clients with the same bitwidth divided by the total number of clients, because the number of clients participating in the aggregation affects the convergence speed of the FL process proportionately~\cite{haddadpour_federated_2020}. We train a weight dequantizer $\phi$ with a SGD optimizer with the learning rate of 0.01, batch size of 16, for 5 epochs for all experiments.

In the experiments in \Cref{tab:main}, we use $\Pi = \{\text{Int2}, \text{Int4}, \text{Int8}, \text{Int16}, \text{Float32}\}$ for the dequantizer blocks, whereas in the experiment in \Cref{tab:multi-scaled-main}, we use $\Pi = \{\text{Int6}, \text{Int8}, \text{Int10}, \text{Int12}, \text{Int16}\}$.

\section{Low-bitwidth Training for Bit-limited Hardware Devices}
\label{appendix:low_bit_traing}

\paragraph{Weight initialization.} In order to prevent weights from vanishing due to the intermediate ternarization, we adjust the scale of the initialization as follows:
\begin{align}
    \bm{w}_q \sim \mathcal{U}(-L, L),
\end{align}
where $L=\max\{0.75, \sqrt{3/\text{fan\_in}_l}\}$. Note that when $L=\sqrt{3/\text{fan\_in}_l}$, it is equivalent to the Kaiming uniform initialization. The layer-wise scaling factor $\alpha^l$ is defined as follows:
\begin{align}
    \alpha^l = \text{Shift}(0.75/\sqrt{3/\text{fan\_in}_l}).
\end{align}
This modified initialization strategy has consequences for the full-precision model, since the scale of the weights aggregated are not the same across bitwidths. We alleviate this problem by initializing the full-precision model weights with the same distribution as the quantized model weights, and using Weight Normalization~\cite{salimans_weight_2016} in the full-precision clients to compensate for the scale difference. % TODO explain modified update scale

\paragraph{Backward pass.} The errors are calculated by using the chain rule, except we normalize the errors at each layer to prevent saturation. Specifically, we apply the following before propagating the error value to every subsequent layer in the chain rule:
\begin{align}
    \bm{e}_q = Q_s(\bm{e}/\mathrm{Shift}(\max\{|\bm{e}|\})),
\end{align}
where $\mathrm{Shift}(x) = 2^{\lceil\log_2 x\rfloor}$ finds the nearest power-of-two to the input, and $\max\{|\bm{e}|\})$ represents the layer-wise maximum absolute value among the elements of the error $\bm{e}$, and $s$ is the bitwidth of the client. The quantizer function is defined as
\begin{align}
    Q_s(x) &:= \text{clip}_s(\lceil(2^{s-1} \cdot x)\rfloor / 2^{s-1}),\\
    \text{clip}_s(x) &:= \max(\min(x, (2^{s-1}-1)/2^{s-1}), (-2^{s-1}+1)/2^{s-1}).
\end{align}

For the weight update, we similarly apply the following rescaling operation to the gradient value before applying the weight update:
\begin{align}
    \bm{q} \leftarrow \mathrm{clip}_{s}(
        \bm{q} - Q_{s}^{\text{stoch}}\left(\eta \cdot \bm{g}/\mathrm{Shift}\left(max\{|\bm{g}|\}\right)\right)
    ),
\end{align}
where $Q_{s_n}^{\text{stoch}}(\cdot)$ is a stochastic quantization function defined elementwise as follows:
\begin{align}
    Q_{s}^{\text{stoch}}(x) = \begin{cases}
        2^{1-{s}} \cdot \lceil|x|\rceil & \text{w.p. } |x|-\lfloor|x|\rfloor \\
        2^{1-{s}} \cdot \lfloor|x|\rfloor & \text{otherwise.}
    \end{cases}
\end{align}

Note that the scale of the learning rate $\eta$ is different from regular full-precision network training, because of the rescaling of the gradient values.

\section{Training of Progressive Weight Dequantizer}
\label{appendix:dequantizer}
\paragraph{Construction of the weights dataset.}
\input{contents/figures/A_weight_dataset}
Given the local model weights $\bm{w}$, we construct the weight datasets to learn the progressive weight dequantizer. Since layers in deep neural networks is often composed of the weights with different dimensionality each other, we split them into the uniformly-sized subweights. As following common structures for CNN models that mostly composed of a number of convolution layers, such as VGG~\cite{simonyan_very_2015} and ResNet~\cite{he2016deep}, we basically utilize convolution weights with a filter size of $3\times3$ and input dimension is $64$ or larger to construct the weight dataset. That is, we use all convolution weights except the weights from the first layer (input dimension is the channel of Image, $3$). We split the weights at each convolution layer to partial modules with the shape of $64\times64\times3\times3$, (e.g., the weights with the shape of $256\times128\times3\times3$ is splitted to $4\cdot2=8$ different modules. Next, we reshape each module sized by $24\times24$ with $64$ channels (i.e., $64\times24\times24$) to as illustrated in \Cref{fig:weight_data}.

\paragraph{Block design for progressive weight dequantizer.} We implement our dimensionality-preserving network block $\phi$ for our dequantizer using two layered of affine coupling layers, and the design of each layer $\rho$ is as follows:
\begin{equation}\label{eq:layer_design}
\begin{split}
\widehat{\bm{w}}_{1:d} &= \bm{w}_{1:d} + \alpha(\bm{w}_{d+1:D}),\\
\widehat{\bm{w}}_{d+1:D} &= \bm{w}_{d+1:D}\odot\exp{\big(\beta(\widehat{\bm{w}}_{1:d})\big)}+\gamma(\widehat{\bm{w}}_{1:d}),\\
\end{split}
\end{equation}
where $\widehat{\bm{w}} = \rho(\bm{w})$, $\alpha$, $\beta$, and $\gamma$ are DenseNet~\cite{iandola2014densenet} blocks while we omit the notation of weights in each layer for readability. To this end, a $j^{th}$ network block of progressive weight dequantizer is formulated as follows:
\begin{equation}\label{eq:block_design}
\begin{split}
\widehat{\bm{w}} = \phi^{\pi_j\rightarrow\pi_{j+1}}(\bm{w}; \bsy\theta_j) = \bm{w} + \tau\rho(\rho(\widehat{\bm{w}})),
\end{split}
\end{equation}
where $\tau$ is a scaling coefficient and we set $\tau=0.1$ for all experiments. We want to note that there is a huge potential to further develop the design of our dequantizer. We intend to suggest a better design for the dequantization function for recovering high-bitwidth weights from low-bitwidth weights in future work.

\section{Additional Experiments}
\label{appendix:add_experiments}

\begin{figure}[th]
    \centering
    \begin{minipage}[c]{\linewidth}
    \vspace{-0.1in}
    \renewcommand{\tabcolsep}{1pt}
    \begin{tabular}{cc} \includegraphics[width=0.95\textwidth]{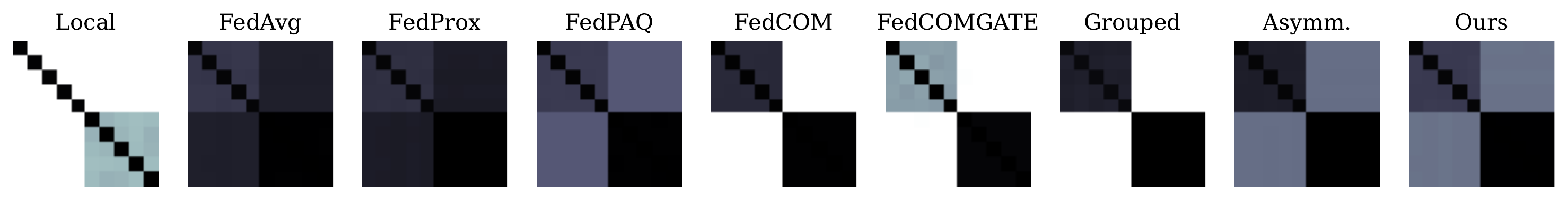}&
    \hspace*{-.1cm}\includegraphics[width=0.044\textwidth]{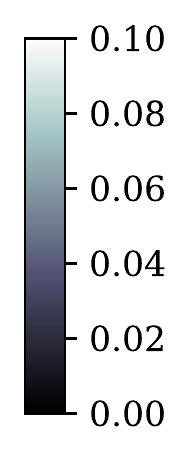}\\
    (a) \textsc{{Int8~(50\%)}~-~{Float32~(50\%)}}\\
    \vspace{-0.1in}\\
    \includegraphics[width=0.95\textwidth]{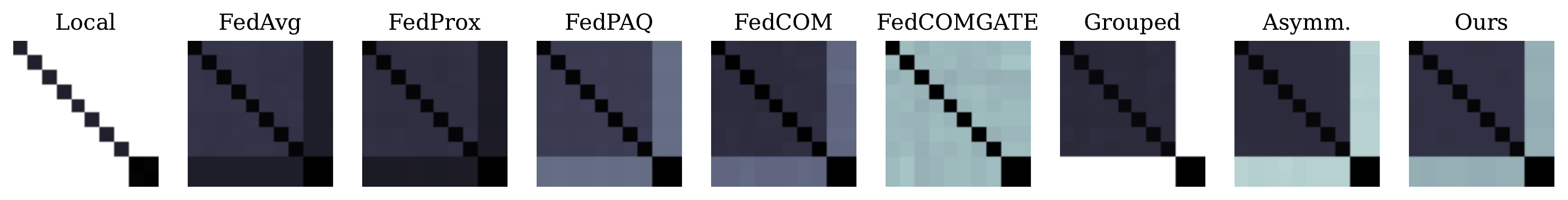}
    &
    \hspace*{-.1cm}\includegraphics[width=0.044\textwidth]{figures/colorbar_2.pdf}
    \\
    (b) \textsc{{Int8~(80\%)}~-~{Float32~(20\%)}}\vspace{-0.05in}
    \end{tabular}
    \end{minipage}
    \caption{\textbf{Cosine distance matrix between the weights of each client.} Elements in a row and column describe the index of local clients. \textbf{(a)} the first five clients correspond to Int8 (Top left) and the other five to Float32 (Bottom right). \textbf{(b)} the first eight clients correspond to Int8, and the other two to Float32. Darker colors indicate higher similarities between clients.}
    \label{fig:dist2}
\end{figure}

\paragraph{Weight similarity anlaysis.}
%\Cref{fig:dist2}~(a) demonstrates the effect of using selective weight aggregation (SGU), which increases the distance between the Int8 and Float32 clients, whereas using a dequantizer only (Deq.~Only) does not. \Cref{fig:dist2}~(b) shows this effect more clearly in an imbalanced client bitwidth setting. We speculate that selective weight aggregation enables the high-bitwidth weights to stay sufficiently far away from the low-bitwidth weights, preventing a distributional shift that degrades their performance. Interestingly, the Int8 clients in FedCOMGATE behave very differently from FedCOM, with the weight distance among them farther than any other methods. This can be explained by the ``correction vectors'' utilized in the FedCOMGATE algorithm that is designed to adapt to data heterogeniety and enable the weights to behave semi-independently. This seems to improve the model accuracy in our main experiment compared to FedCOM as seen in \Cref{tab:main}, but only by a marginal amount. In contrast, our selective weight aggregation is better suited for BHFL for preserving the models weights' heterogeneity.

We further provide the cosine distance matrix between the weights of each client for all baselines in \Cref{fig:dist2}. Each client in the local training (Local) model does not share the knowledge with other clients, resulting in a large weight distance. FedProx~\cite{li2018federated} shows similar tendency with FedAvg. Interestingly, the Int8 clients in FedCOMGATE show different behaviors with FedCOM, with the weight distance among them farther than any other BHFL methods. This phenomenon is due to the ``correction vectors'' utilized in the FedCOMGATE algorithm designed to adapt to data heterogeniety and enable the weights to behave semi-independently. This seems to improve the model accuracy compared to FedCOM with a marginal amount, as shown in \Cref{tab:main}. While FedGroupedAvg-Asymmetric (Asym.) shows a similar distance matrix with FedGroupedAvg (Grouped), it allows the knowledge transfer from-high-to-low-bitwidths clients, performing higher weight similarity between bitwidth heterogeneous clients.

\begin{table*}[h]
% \vspace{-0.05in}
\caption{\small Average accuracy at each bitwidth and average accueacy across all clients on CIFAR-10 dataset. We set participating clients with 50\% of Int8 and 50\% of Float32 models (Left), and 50\% of Int8 and 50\% of Float32 models (Right). All of the results are measured by computing the 95\% confidence interval over three independent runs.}
\label{tab:main}
\vspace{-0.05in}
\centering
\small
\setlength{\tabcolsep}{3pt} % 
\resizebox{0.95\textwidth}{!}{
\begin{tabular}{l@{\hspace{6pt}}c c  a c c c a}
%\begin{tabular}{llcccccc}
\toprule
{\textsc{Method}}&\multicolumn{3}{c}{\textbf{\textsc{CIFAR-10}, {Int8~(50\%)~-~{Float{32}~(50\%)}}}} &~& \multicolumn{3}{c}{\textbf{\textsc{\textbf{CIFAR-10}, {Int8~(80\%)}~-~{Float{32}~(20\%)}}}}\\
\midrule
%& & Accuracy ($F$) & Accuracy ($I$) & Averaged & Accuracy ($F$) & Accuracy ($I$) & Averaged\\
%& & $F32$ Accuracy & $I8$ Accuracy & Average & $F32$ Accuracy & $I8$ Accuracy & Average\\
& \textsc{int8 acc} & \textsc{float32 acc} &\textsc{average} && \textsc{int8 acc} & \textsc{float32 acc} & \textsc{average}\\
\midrule
%\parbox[t]{2mm}{\multirow{7}{*}{\rotatebox[origin=c]{90}{\small  $\bm{F{32}~(5)}~-~\bm{I8~(5)}$}}}
%\parbox[t]{2mm}{\multirow{9}{*}{\rotatebox[origin=c]{90}{\small  \textsc{{F{32}~(50\%)}~-~{Int8~(50\%)}}}}}
\textsc{Local Training} & 
{69.59}~\scriptsize($\pm$~0.34) &
{75.82}~\scriptsize($\pm$~0.41) &
{72.71}~\scriptsize($\pm$~0.26) &&
{69.39}~\scriptsize($\pm$~0.34) &
{76.41}~\scriptsize($\pm$~0.52) &
{70.79}~\scriptsize($\pm$~0.34)\\
\midrule
\textsc{FedAvg}& 
{76.88}~\scriptsize($\pm$~0.49) &
{76.23}~\scriptsize($\pm$~0.36) &
 {76.56}~\scriptsize($\pm$~0.36) &&
{77.43}~\scriptsize($\pm$~0.83) &
{74.64}~\scriptsize($\pm$~1.21) &
{76.87}~\scriptsize($\pm$~0.87)\\
\textsc{FedProx} & 
{71.16}~\scriptsize($\pm$~0.35) &
{69.28}~\scriptsize($\pm$~0.90) &
{70.22}~\scriptsize($\pm$~0.55) &&
{69.60}~\scriptsize($\pm$~0.50) &
{66.28}~\scriptsize($\pm$~1.24) &
{68.94}~\scriptsize($\pm$~0.57)\\
\textsc{FedCOM}& 
{75.37}~\scriptsize($\pm$~0.52) &
{77.69}~\scriptsize($\pm$~0.45) &
{76.53}~\scriptsize($\pm$~0.38) &&
{73.60}~\scriptsize($\pm$~1.08) &
{80.73}~\scriptsize($\pm$~0.73) &
{75.03}~\scriptsize($\pm$~0.94)\\

\textsc{FedGroupedAvg} & 
{61.85}~\scriptsize($\pm$~0.78) &
{85.08}~\scriptsize($\pm$~0.28) &
{73.46}~\scriptsize($\pm$~0.29)&&
{71.76}~\scriptsize($\pm$~0.68) &
{78.07}~\scriptsize($\pm$~0.58) &
{73.02}~\scriptsize($\pm$~0.65) \\

\midrule
%& \textsc{HIVE w/o dequantizer (Ours)} & 
%{79.79} \scriptsize($\pm$ 0.98) & \textbf{85.94} \scriptsize($\pm$ 0.92) & {82.84} %\scriptsize($\pm$ 0.55) &&
%{*} \scriptsize($\pm$ *) & {*} \scriptsize($\pm$ *) & {*} \scriptsize($\pm$ *) \\
\textsc{ProWD (Ours)} & 
\textbf{82.87}~\scriptsize($\pm$~0.37) &
\textbf{85.99}~\scriptsize($\pm$~0.20) &
\textbf{84.43}~\scriptsize($\pm$~0.22)&&
\textbf{79.23} \scriptsize($\pm$~0.25) &
{81.26} \scriptsize($\pm$~0.45) &
\textbf{79.63} \scriptsize($\pm$~0.23)\\
\bottomrule
\end{tabular}}
\end{table*}

%% file: contents/figures/A_weight_dataset.tex
\begin{wrapfigure}{h!}{0.35\textwidth}
\centering
\vspace{-0.45in}
\includegraphics[width=0.35\textwidth]{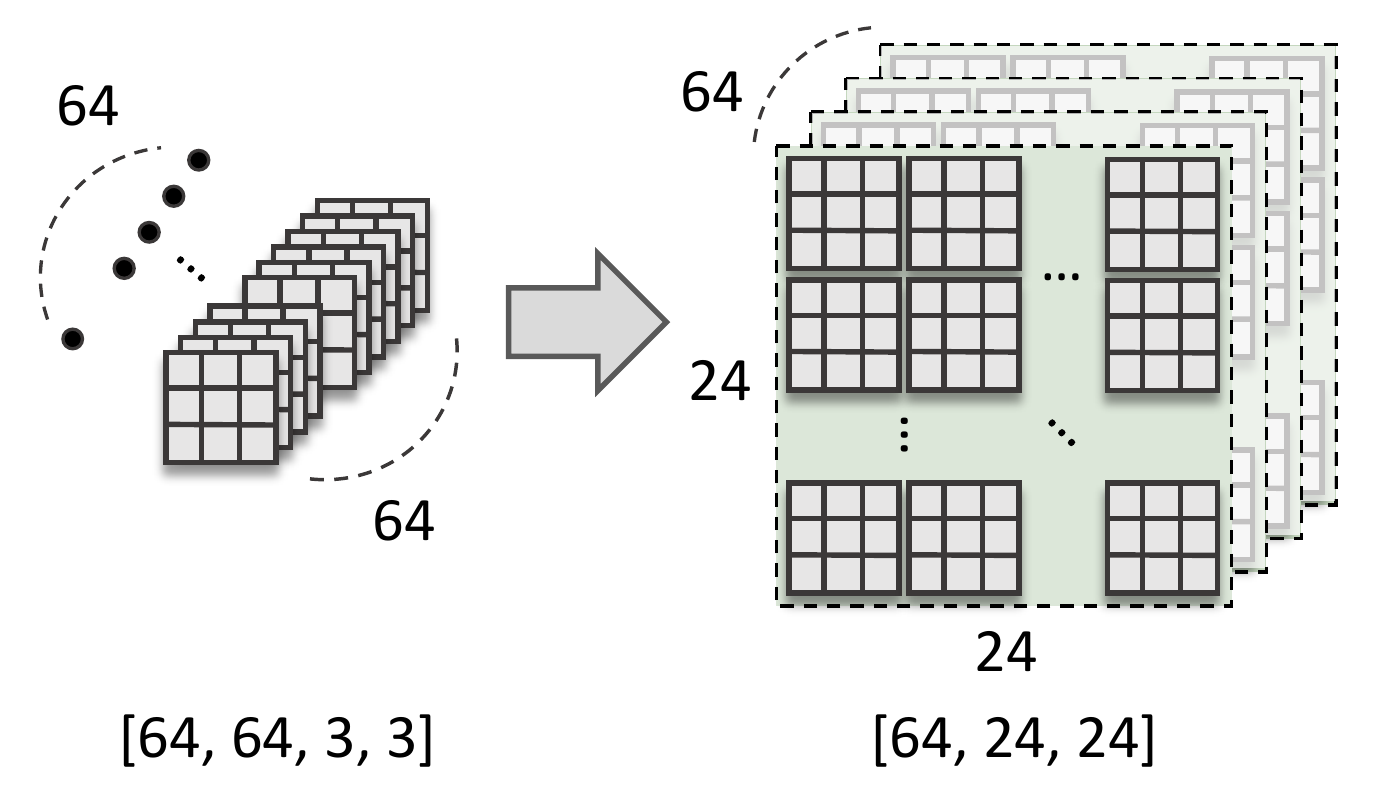}
\vspace{-0.3in}
\caption{\label{fig:weight_data} \small \textbf{Illustration of a reshaping process on the weight module.}}
\vspace{-0.1in}
\end{wrapfigure}